\documentclass[11pt]{article}
 \usepackage[total={6in, 8in}]{geometry}
 \usepackage{adjustbox}

\usepackage[utf8]{inputenc} 
\usepackage[T1]{fontenc}    
\usepackage{url}            
\usepackage{booktabs}       
\usepackage{amsfonts}       
\usepackage{nicefrac}       
\usepackage{microtype}      
\usepackage{xcolor}         
\usepackage{array}
\usepackage{makecell}
\usepackage{tabularray}
\UseTblrLibrary{booktabs}

\usepackage{natbib}

\usepackage[colorlinks,linkcolor=red,citecolor=blue, pagebackref=true]{hyperref}

\usepackage{amsmath,amssymb,amsthm}
\usepackage{tikz}
\usepackage{comment}
\usepackage{color}
\usepackage{setspace}
\usepackage{pdflscape}
\usepackage{enumitem}
\usepackage{algorithm}
\usepackage{algpseudocode}
\usepackage{dsfont}

\usepackage[english]{babel}
\usepackage{macros}

\makeatletter
\renewcommand\tableofcontents{%
    \@starttoc{toc}%
}
\makeatother

\usepackage{amsmath}
\makeatletter
\let\save@mathaccent\mathaccent
\newcommand*\if@single[3]{%
  \setbox0\hbox{${\mathaccent"0362{#1}}^H$}%
  \setbox2\hbox{${\mathaccent"0362{\kern0pt#1}}^H$}%
  \ifdim\ht0=\ht2 #3\else #2\fi
  }
\newcommand*\rel@kern[1]{\kern#1\dimexpr\macc@kerna}
\newcommand*\widebar[1]{\@ifnextchar^{{\wide@bar{#1}{0}}}{\wide@bar{#1}{1}}}
\newcommand*\wide@bar[2]{\if@single{#1}{\wide@bar@{#1}{#2}{1}}{\wide@bar@{#1}{#2}{2}}}
\newcommand*\wide@bar@[3]{%
  \begingroup
  \def\mathaccent##1##2{%
    \let\mathaccent\save@mathaccent
    \if#32 \let\macc@nucleus\first@char \fi
    \setbox\z@\hbox{$\macc@style{\macc@nucleus}_{}$}%
    \setbox\tw@\hbox{$\macc@style{\macc@nucleus}{}_{}$}%
    \dimen@\wd\tw@
    \advance\dimen@-\wd\z@
    \divide\dimen@ 3
    \@tempdima\wd\tw@
    \advance\@tempdima-\scriptspace
    \divide\@tempdima 10
    \advance\dimen@-\@tempdima
    \ifdim\dimen@>\z@ \dimen@0pt\fi
    \rel@kern{0.6}\kern-\dimen@
    \if#31
      \overline{\rel@kern{-0.6}\kern\dimen@\macc@nucleus\rel@kern{0.4}\kern\dimen@}%
      \advance\dimen@0.4\dimexpr\macc@kerna
      \let\final@kern#2%
      \ifdim\dimen@<\z@ \let\final@kern1\fi
      \if\final@kern1 \kern-\dimen@\fi
    \else
      \overline{\rel@kern{-0.6}\kern\dimen@#1}%
    \fi
  }%
  \macc@depth\@ne
  \let\math@bgroup\@empty \let\math@egroup\macc@set@skewchar
  \mathsurround\z@ \frozen@everymath{\mathgroup\macc@group\relax}%
  \macc@set@skewchar\relax
  \let\mathaccentV\macc@nested@a
  \if#31
    \macc@nested@a\relax111{#1}%
  \else
    \def\gobble@till@marker##1\endmarker{}%
    \futurelet\first@char\gobble@till@marker#1\endmarker
    \ifcat\noexpand\first@char A\else
      \def\first@char{}%
    \fi
    \macc@nested@a\relax111{\first@char}%
  \fi
  \endgroup
}
\makeatother

\usepackage{mathtools}

\usepackage{caption}
\usepackage{subcaption}

\usepackage{booktabs}

\usepackage{nameref}
\usepackage[capitalize]{cleveref}

\usepackage[T1]{fontenc}
\usepackage{underscore}

\usepackage{listings}
\usepackage[dvipsnames]{xcolor} 

\Crefname{theorem}{Theorem}{Theorems}
\Crefname{lemma}{Lemma}{Lemmas}
\title{Fidelity-Diversity Metrics for Text}

\author{
Amanda Wang\footnote{Department of Computer Science, Cornell University}
\quad
Tudor Manole\footnote{Statistics and Data Science Center, Massachusetts Institute of Technology}
\quad
Florentina Bunea\footnote{
Department of Statistics, Cornell University}
\quad
John Thickstun$^{*}$\\[.5ex]
\footnotesize
\texttt{arw274@cornell.edu, tmanole@mit.edu, fb238@cornell.edu, jthickstun@cornell.edu}
}

\date{}

\begin{document}

\maketitle

\begin{abstract} 
As language modeling technology matures, there is an increasing research focus on the composition and curation of datasets used to train these models. For instance, practitioners commonly seek to augment high-quality datasets with additional text to enhance the performance of models trained on that data. However, informed decisions about data augmentation require more nuanced assessments about data quality.
We build on work measuring the precision and recall of generative models to develop a pair of  metrics that quantify (1) fidelity, capturing how closely candidate text resembles reference data, and (2) diversity, capturing how well it covers the modes of the reference dataset. Our metrics are 
based on optimal transport divergence functionals
between discrete text summaries. In experiments on M2D2 text datasets, we show that these metrics are able to disentangle a lack of fidelity from a lack of diversity in deficient candidate text. In further experiments, our metrics detect diversity deficits in synthetic GSM8K-style math datasets, which correlate with degradations in downstream accuracy of language models finetuned on this synthetic data.
\end{abstract}

\section{Introduction}
 
In tandem with the development of large language models (LMs), the research community has developed quantitative evaluation metrics to assess the quality of text generated by these models. In this paper we take a broader perspective on the evaluation of text, reimagining this style of quantitative evaluation as a tool to improve our understanding of the datasets used for training and finetuning LMs. We are particularly interested in synthetic training data, which bridges model evaluation and dataset analysis: in this setting, generated text is both the training data input and sampled output of language models \cite{gunasekar2023textbooks,wang2023self}.

The predominant metrics for evaluating LM-generated text compare model outputs to reference text using a scalar numerical score. These metrics can be computed either on the instance level, comparing one model output to one reference~\cite{zhangbertscore}, or between i.i.d. samples from distributions of generated text and reference text~\cite{pillutla2021mauve}. We study the latter scenario, comparing an \emph{evaluation dataset} collected from one distribution of text (possibly LM-generated, but not necessarily) to a \emph{reference dataset}. We seek to develop more nuanced metrics for insight into data quality, disaggregating our measure of data quality along two axes: \emph{fidelity}, how similar evaluation text is to the reference text, and \emph{diversity}, how broadly the evaluation text covers the modes of the reference text distribution.

We quantify fidelity and diversity through two separate notions of transport between discrete measures. Our metrics operate on pretrained text embeddings; each dataset is summarized by a discrete measure consisting of point masses on a finite set of embeddings that represent modes of the data distribution. Fidelity is defined through the cost of a greedy nearest neighbor assignment between evaluation modes and reference modes. We arrive at this notion through an optimal transport argument, providing an alternative motivation for nearest-neighbor metrics that have previously proposed as a measure of data fidelity~\citep{sajjadi2018assessing}.
Diversity is defined as the {\it gap} between this greedy
assignment and the global cost of optimal transport between the two measures. Taken together, these metrics decompose the Wasserstein transport cost between dataset representations into two components: costs arising from lack of fidelity (geometric mismatch in support) and costs arising from lack of diversity (discrepancy of mass distribution).






We study the behavior of these metrics in two settings. First, we seek to validate their efficacy 
in a setting with ground-truth labels that reflect a colloquial understanding of fidelity and diversity. Using text from the M2D2 dataset~\citep{reid2022m2d2} annotated with topic labels, we design an experiment that controls the topics contained in the evaluation dataset and measures the sensitivity of our metrics as we vary the overlap and coverage of topics relative to the reference dataset. Second, we demonstrate that our metrics are sensitive to more subtle variations in fidelity and diversity, in a setting where no clear topic structure exists. We design a protocol for generating synthetic variants of the GSM8K dataset~\citep{cobbe2021training} that allows us to control for the diversity of problems contained in each generated variant. We show that our diversity metric is sensitive to degraded diversity in these datasets and furthermore, this diversity score is predictive of downstream GSM8K accuracy when we finetune LMs on these synthetic datasets.

Our main contributions are summarized as follows. First, we propose principled fidelity-diversity metrics based on a decomposition of the Wasserstein transport distance between distributions. Second, we devise a controlled experiment to validate that these metrics behave intuitively based on data with ground-truth topic labels. Third, we devise a protocol for controlling diversity when constructing synthetic datasets. Using this protocol, we construct synthetic variants of the GSM8K dataset with varying degrees of diversity; we show that our diversity metric is sensitive to the diversity in these datasets, and that diversity in these synthetic training datasets directly correlates with downstream accuracy of models trained on this data. 

\section{Related work}

\paragraph{Evaluating generative models.}

There is a   large body of prior work that has sought
to evaluate generative models through distributional
divergence measures. Widely-used examples
include the Inception Score~\citep{salimans2016improved}
or the more recent Fr\'echet Inception Distance~(FID; \citep{heusel2017}), 
which approximates both distributions in 
a learned feature space by multivariate Gaussian distributions, comparing their means and covariances using
the Wasserstein distance.  
However, global metrics like FID
provide a single-number summary of 
a generative model's behavior, and cannot
distinguish between different failure modes. 

In contrast, a separate line of work
has sought to disentangle different
aspects of a generative model's quality through
the lens of {\it precision} and {\it recall}
metrics. 
This perspective
was introduced through the work of~\citep{sajjadi2018assessing}, among others, 
and was subsequently developed
into a variety of practical methodologies for evaluating generative models~\citep{kynkaanniemi2019improved,naeem2020reliable,alaa2022faithful,jiralerspong2023feature,djolonga2020precision}.
While these methods have
been successful for applications
involving image generation, 
they have seen more limited
use in the text domain. 

The works~\cite{pillutla2023mauvescoresgenerativemodels,pillutla2021mauve} 
introduced the MAUVE score for evaluating text data, 
which sits in between global divergence metrics like FID, 
and precision-recall based methods. Rather than isolating
fidelity and diversity as individual metrics, the MAUVE score 
reports a divergence frontier, which roughly represents
the extent to which improvements of one deteriorate the other.

Closest to our work is~\cite{le2024exploring}, which
develops precision-recall metrics specifically for 
evaluating language models. Their framework adapts the metrics of~\cite{kynkaanniemi2019improved}
to the text domain, and empirically demonstrates their efficacy
in large-scale open-ended text generation tasks. 
Our work takes a complementary perspective to theirs. The focus of our work is not on LM evaluation, but rather
on leveraging fidelity-diversity metrics to assess datasets used for training and finetuning LMs. 
We provide further discussion of methodological differences between our fidelity-diveristy metrics and their precision-recall metrics in Appendix \ref{app:pr}, as well as an empirical comparison of our methods on Wikipedia text data.

Finally, it is also
possible to measure diversity without relying on a reference
corpus, but to instead quantify intrinsic properties
of generated text~\citep{friedmanvendi}. We refer to~\cite{celikyilmaz2020evaluation,ito2025reference} 
for  surveys of such methods.



\paragraph{Optimal transport for text comparison.}
The Wasserstein distance between
mixing measures
has previously been adopted
for comparing text corpora
in~\cite{BBJ25,Bing2022}, and has more broadly
been advocated as a loss function in latent variable
models at least since the works~\cite{nguyen2013,nguyen2015}.
In particular, Ref.~\cite[Theorem 1]{BBJ25} shows that
the Wasserstein distance is 
the most discriminative metric between mixing measures
that is jointly convex and consistent with the ground metric over mixture
components. We view this result as strong motivation for
adopting the Wasserstein distance in our work.
When one relaxes the metric requirement, stronger divergences
have also recently been considered~\citep{manole2022refined,nguyen2024sigmoid}. 
From a different perspective, optimal transport  
has been used
to compare text documents
without quantization~\cite{kusner2015word,yurochkin2019hierarchical,alvarez2020geometric}.
In either case, we are not aware of past attempts
at using it to disentangle notions of fidelity and diversity. 
Finally, we mention a separate line of
work, in which optimal transport has
been used as a means
of aggregating 
quality metrics for language
models using tests
for stochastic dominance~\cite{rioux2024multivariate}.

\section{Methodology} 

\subsection{Problem setting}

 Let $\calT$ be a finite vocabulary of tokens, and let
$\calT^*$ denote the set
of all sequences of tokens
of arbitrary, finite,  length. Assume that we
are given two text datasets of the form
$$ \widebar \calU=\big\{\widebar U_1,\dots,\widebar U_n\big\} \subseteq \calT^*, \quad  \widebar \calV = \big\{\widebar V_1,\dots,\widebar V_m\big\} \subseteq \calT^*.$$
We interpret $\widebar \calU$
as an {\it evaluation dataset}---a collection
of text documents whose fidelity and diversity 
we wish to evaluate against
the {\it reference dataset} $\widebar \calV$. 
Our aim 
is to define metrics for
quantifying the quality
of the evaluation corpus that explicitly
distinguish between its fidelity
and diversity.

Irrespective of the downstream application,
our metric construction  will rely on 
 first embedding the 
two datasets into a common 
  space $\bbR^L$, where 
  the dimension $L\geq 1$
  does not depend
  on the lengths of
  the various documents. 
Concretely, assume that we have
access
an embedding model $\calE:\calT^*\to \bbR^L$,
such as a Sentence Transformer~\cite{reimers2019sentence},
and define the embedded documents
$U_i = \calE(\widebar U_i)$ and $V_j = \calE(\widebar V_j),$
to obtain  the embedded datasets 
$$  \calU=\big\{ U_1,\dots, U_n\big\} \subseteq \bbR^L, \quad   \calV = \big\{V_1,\dots,V_m\big\} \subseteq \bbR^L.$$

In order to motivate
the definition of our metrics, 
we will treat each corpus as a collection of
i.i.d.~draws from probability
distributions over $\bbR^L$, 
\begin{align} 
\label{eq:datasets} 
U_1,\dots,U_n\sim P, \quad V_1,\dots,V_m \sim Q.
\end{align}
Our fidelity and diversity
metrics will be based on 
discrete representations of these 
probability distributions, described next, which will
be estimated on the basis of the
two datasets. 
 
 

\subsection{Discrete measure representations} 

{\bf Quantization.}
We seek  discrete  summaries
of the  distributions $P$ and~$Q$
in the form of {\it mixing measures}: 
discrete probability
distributions which respectively
take the following form:
$$G  = \sum_{i=1}^k \alpha_i
\delta_{\theta_i},
\quad H = \sum_{i=1}^k \beta_i
\delta_{\eta_i},$$
where  $\alpha=(\alpha_1,\dots,\alpha_k)$
and $\beta=(\beta_1,\dots,\beta_k)$
are elements of the $(k-1)$-dimensional
simplex~$\Delta_k$, 
and $\theta_1,\eta_1,\dots,\theta_k,\eta_k$
lie in a parameter space $\Theta \subseteq \bbR^L$ 
for some $k\geq 1$. Here,~$\delta_\theta$ denotes the Dirac
mass (degenerate probability
distribution) at $\theta$.
 By potentially setting some  of the mixing proportions 
$\alpha_i$ and $\beta_i$
to zero, we can assume
without loss of generality that
both mixing measures have the same number
of components $k$.
We will interpret the
{atoms} $\theta_i$ and $\eta_i$
as $k$ latent semantic
topics that are captured
by the evaluation and reference datasets, respectively. 
In turn, the mixing probabilities~$\alpha_i$ and~$\beta_i$
denote the relative weight assigned
to topic $i$
under each corpus.

There are many  ways of 
constructing such quantizations. Perhaps the most natural
approach is to approximate
$P$ and $Q$ with mixture models. We assume, for simplicity
of exposition,  that $P$ and $Q$ admit respective densities
$p$ and $q$.  Concretely, given a parametric
family of densities $F=\{f_\theta:\theta\in\Theta\}$
over $\bbR^L$, we say that $P$ is a mixture
model with mixing measure $G$ if its density
takes the form
\begin{align}\label{eq:mixture}
p(x) = \int_\Theta f_\theta(x)dG(\theta) = \sum_{i=1}^k \alpha_i f(x;\theta_i),\quad x \in \bbR^L.
\end{align}
For instance, the class $F$ could be the set
of Gaussian densities $N(\mu,\Sigma)$, with
respect to the location-scale parameter
$\theta=(\mu,\Sigma)$,
or a kernelized Gaussian mixture family
with a learned kernel.  


%

In practice, the datasets~\eqref{eq:datasets} are used
to learn estimators
$\hat G = \sum_{i = 1}^{k}  \hat \alpha_i
\delta_{ \hat \theta_i},  \  \hat H = \sum_{i=1}^k \hat \beta_i
\delta_{\hat \eta_i}$
of the mixing measures $G$ and $H$, giving rise to the data-dependent quantizations of a corpus. 
For instance, if quantization arises via a mixture approximation, one can
use the Expectation-Maximization 
algorithm to approximate the maximum likelihood
estimator under model~\eqref{eq:mixture}, which we shall
pursue in Section~\ref{M2D2}. 
More generally, many clustering
procedures can be understood as producing
discrete
mixing measure summaries such as $\hat G$ and $\hat H$, and therefore fit naturally
within this framework. 



{\bf Empirical measures.} There may be situations in which the data does not naturally contain interpretable  semantic topics, in which case  quantization becomes much less informative.  In that case,  we can use the empirical measures
$\frac{1}{n}\sum_{i = 1}^{n}\delta_{U_{i}}$ and $\frac{1}{m}\sum_{i=1}^{m}\delta_{V_{i}}$ as canonical data-dependent discrete representations of $P$ and $Q$.  We illustrate this  in Section \ref{Syn}.

We emphasize that
the metrics we propose in the next section are independent of the way in which we arrive at discrete  representations $G$ and $H$ of the distributions
$P$ and $Q$.


\subsection{Optimal transport fidelity and diversity metrics  }\label{metrics}

We are now ready to introduce
fidelity  and diversity metrics
for comparing the evaluation distribution
$P$ to the reference distribution $Q$,
via their mixing measures $G$ and $H$, respectively.
Our metrics
are based on the {\it optimal transport}
problem~\citep{villani2003topics,peyre2019computational,chewi2025statistical} between $G$ and~$H$, which
we recall next. 
\begin{definition}
Given two mixing measures 
$G = \sum_{i=1}^k\alpha_i\delta_{\theta_i}$
and $H=\sum_{i=1}^k \beta_i\delta_{\eta_i}$, 
we denote by $\Pi(G,H)$ 
the set of doubly stochastic matrices
which are {\it couplings} of $G$
and $H$:
$$\Pi(G,H) = 
\Big\{\gamma \in \bbR_+^{k\times k}:  
\textstyle\sum_{i'=1}^k \gamma_{i'j}= \beta_j, 
\sum_{j'=1}^k \gamma_{ij'} = \alpha_i,\text{for all } i,j\in \{1,\dots,k\}\Big\}.$$
Given a metric $d(\cdot,\cdot)$ on $\Theta$, the 1-Wasserstein distance between
$G$ and $H$ is defined by
$$W_1(G,H) = \inf_{\gamma\in \Pi(G,H)} \sum_{i,j=1}^k \gamma_{ij} d(\theta_i,\eta_j),$$
and any solution to the above
optimization problem
is called an optimal transport coupling.

\end{definition}

The Wasserstein distance  measures the most efficient way of redistributing
the probability mass of 
$G$ onto that of $H$,
as measured by the metric $d$ on the underlying
space. 
In doing so, it induces a 
global notion of 
distance between the 
evaluation distribution $P$ 
and the reference distribution $Q$. 
Our goal will now be to decompose
the  distance $W_1(G,H)$ into
two complementary  metrics 
which capture
two distinct aspects of 
the evaluation distribution:  the fidelity and the diversity of $G$ relative to $H$. 

\begin{figure}[t!]
\centering
\includegraphics[width=\textwidth]{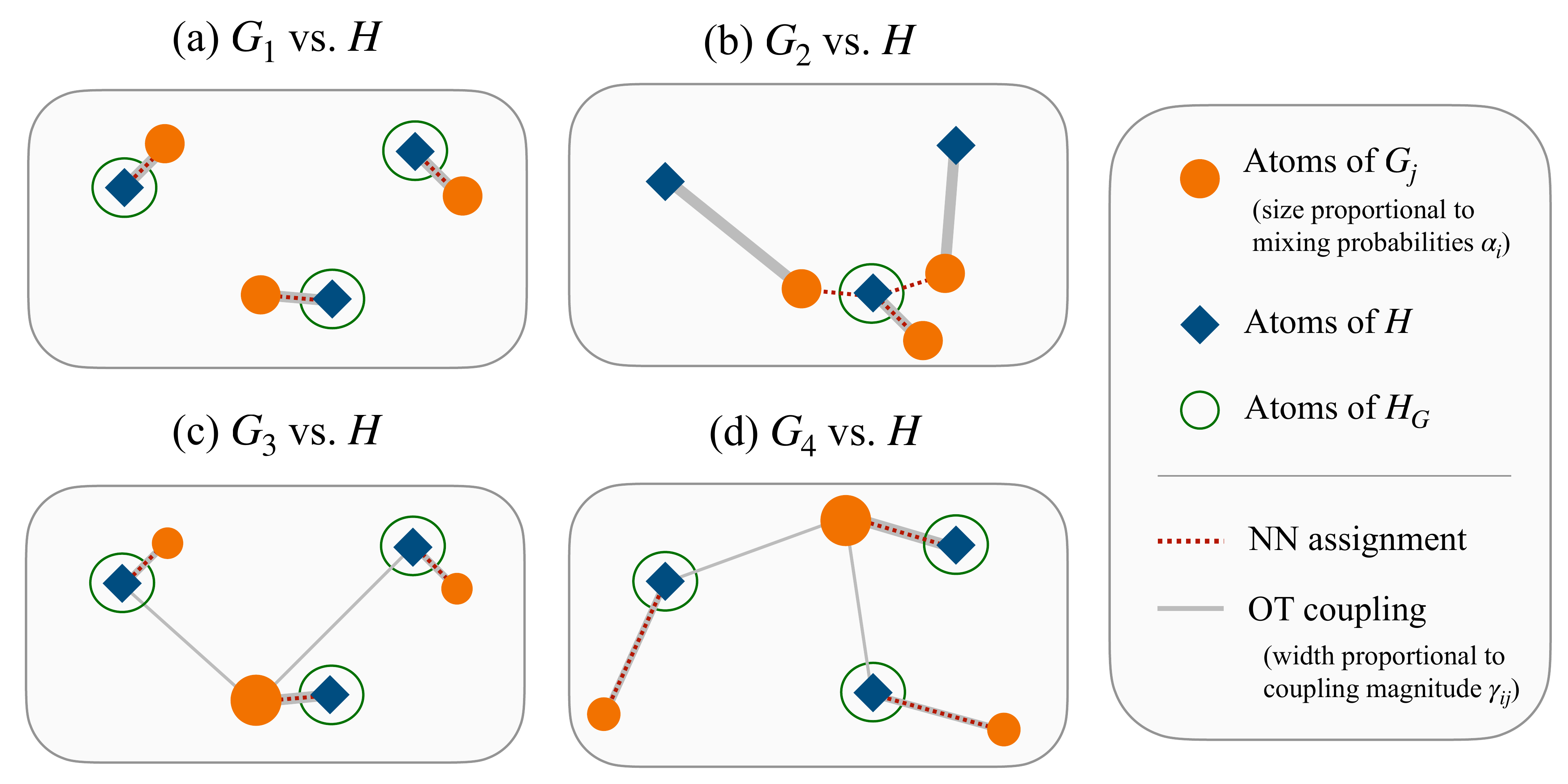}
\caption{ 
 Illustration of four evaluation
 mixing measures $G_1,\dots,G_4$ (with atoms in orange), 
 and one reference mixing measure $H$ (with atoms in blue),
 lying in dimension $L=2$. The dotted
 lines denote the nearest blue point to each orange point,
 whereas the grey lines denote the magnitude of the optimal
 coupling between the measures $G_i$ and $H$. 
 In each figure, the fidelity $\calF_i=\calF(G_i,H)$ 
 can be read as a weighted average
 of the length of the dotted lines, whereas
 the diversity $\calD_i=\calD(G_i,H)$ can be read
 as the difference between  weighted averages
 of the grey and dotted lines. The top row
 illustrates settings where all mixing probabilities
 are equal, and the optimal coupling is a perfect matching, whereas the bottom
 row illustrates settings where 
 the mixing probabilities are not uniform. Figure (a)
 illustrates high fidelity and perfect diversity. 
 Figure (b) illustrates high fidelity and low diversity. 
 Figure (c) illustrates high fidelity
 and fair diversity. Figure (d) illustrates low fidelity
 and fair diversity.  
 In summary,  we have 
 $\calF_1= \calF_2=\calF_3  \geq \calF_4$,
 and $1=\calD_1\geq\calD_3= \calD_4 \geq \calD_2$. 
\label{fig:metrics}
}
\end{figure}

\paragraph{Fidelity metric.}
To isolate
how well the support
of $G$ aligns with that
of $H$, we will define a third
mixing measure $H_G$, 
defined as  the best
approximation of $G$
using {\it only} atoms from~$H$.
That is, letting
$\calP(H)$ denote the set
of mixing measures whose
support is contained
in that of $H$, we define
$$H_G = \argmin_{\mu\in \calP(H)}
W_1(G,\mu),$$
assuming that a minimizer
exists.  
Intuitively, $H_G$ is obtained
by reassigning all atoms of~$G$
onto the atoms of $H$, without
requiring that all atoms of $H$ receive mass.  
We then define
the {\it fidelity} of $G$
relative to $H$ as follows:
$$\calF(G,H) = \exp\left\{-\frac{W_1(G,H_G)}{\kappa(H)}\right\},
\quad \text{with }
\kappa(H) := \bbE_{\eta,\eta'\sim H} [d(\eta,\eta')]=\sum_{i,j=1}^k \beta_i\beta_j d(\eta_i,\eta_j).
$$
 
A simple derivation (cf.~App.~\ref{app:lemma})
shows that 
the Wasserstein distance 
between $G$ and $H_G$ is available
in closed-form, and arises
from a nearest-neighbor search:
One has the equivalent expression
\begin{align}\label{eq:nn_identity}
\calF(G,H) = \exp\left\{-\frac 1 {\kappa(H)}\sum_{i=1}^k \alpha_i \cdot \min_{1 \leq j \leq k} d(\theta_i,\eta_j)\right\}.
\end{align}
The fidelity thus
measures
how close the atoms of $G$
are to {\it some} atom of $H$. It
is large when each
atom of $G$ lies near an atom of $H$,
indicating heuristically that each
semantic topic captured by the evaluation corpus comes close
to some topic in the reference corpus; see Figure\,\ref{fig:metrics}(a) for an illustration.
However, the fidelity
can still be large if 
{\it some}  of the atoms
of $H$ are far from those of $G$,
which reflects a lack
of diversity rather than fidelity;  see Figure\,\ref{fig:metrics}(b).
In this case, all topics in the  evaluation corpus $G$ still come close  to the reference corpus $H$,  but this is achieved  by being  faithful to only one  sematic topic in $H$.

We  include the normalization $\kappa(H)$ so that the fidelity
can be interpreted on an absolute scale. 
In particular, we  expect a mixing measure $G$ that has high fidelity   to have its atoms located closer
to those of $H$ than the typical
 distance between the atoms of $H$ themselves. 
 One can thus loosely
 think of $e^{-1}$ as a lower bound on 
 typical values of $\calF(G,H)$ for high-fidelity
  text.

\paragraph{Diversity metric.}
We further define
the {\it diversity} of $G$
relative to $H$ as
$$\calD(G,H) = \exp\left\{- \frac{1}{\kappa(H)} \Big(W_1(G,H) - W_1(G,H_G)\Big)\right\}.$$
This quantity captures
the extent to which  $G$  spreads
its mass
around the support of $H$. 
To make this interpretation clear, we consider
the special case where 
the mixing measures are uniform: 
$\alpha_i = \beta_i = 1/k$ for all $i=1,\dots,k$. 
In this case, the optimal transport
coupling is always achieved
by a permutation: Denoting by
$\calS_k$ the permutation group on $k$ elements, 
one has
\begin{align*} 
\calD(G,H) = \exp\left\{-\frac {\Delta} {\kappa(H)}\right\}, \quad \text{with }\Delta
 &= \frac 1 k \min_{\sigma\in \calS_k} 
 \sum_{i=1}^k \Big\{d(\theta_i,\eta_{\sigma(i)})
  - \min_{1 \leq j \leq k} d(\theta_i,\eta_j)\Big\}.
 \end{align*}
 The quantity $\Delta$ measures the difference between the cost of the
 greedy nearest neighbor matching
 from the atoms of $G$ to those of $H$, 
 and the optimal transport cost, which must
 respect the constraint of mapping
   mass from $G$ onto all atoms of $H$. 
 Intuitively, if $G$ has high diversity, 
 it should spread its mass evenly
 around the atoms of $H$, so these
 two costs should be similar, and $\Delta$
 should be small.  
 

\paragraph{Fidelity-diversity metric decomposition.}
Taken together, our
definitions of fidelity and
diversity induce
a decomposition 
$$
\exp\left(-\frac{1}{\kappa(H)} W_1(G,H)\right) = \calF(G,H)\cdot \calD(G,H).$$ 
of the total Wasserstein distance
between $G$ and $H$,
suitably scaled and exponentiated, 
into its contributions
arising from lack
of fidelity (geometric mismatch
in support), and diversity (discrepancy in mass distribution). 
We refer to Figure~\ref{fig:metrics}
for an illustration of the behavior
of our metrics on several examples.

\section{Experiments}
\label{sec:experiment}
We conduct two sets of experiments. The first seeks to demonstrate the efficacy of our method for measuring diversity in text by validating our method against ground-truth categorizations of text data from the M2D2 corpus~\cite{reid2022m2d2}. We find that our method produces measurements which correlate well with the amount of diversity implied by these categorizations. The second experiment seeks to demonstrate the usefulness of our method for assessing diversity in synthetic training data. More specifically, we devise a controlled procedure for generating multiple synthetic GSM8K-style math datasets of increasing diversity. We then use our metrics to assess their fidelity and diversity against a reference dataset, and find that our measurements align well with the experimental design: datasets that are constructed to be more diverse receive higher diversity scores, while all achieve similar fidelity scores, as we would expect based on the synthesis procedure. Finetuning a language model on each synthetic dataset further demonstrates that our diversity measurements correlate well with downstream model performance when evaluated on a held-out test set. We open-source our code\footnote{\href{https://github.com/arw274/fidelity-diversity}{https://github.com/arw274/fidelity-diversity}} and synthetic datasets\footnote{\href{https://huggingface.co/datasets/arw274/gsm8k-synthetic}{https://huggingface.co/datasets/arw274/gsm8k-synthetic}}. 

\subsection{M2D2 experiments}\label{M2D2}
We perform several controlled experiments using human-annotated text data to empirically validate whether our metrics successfully disaggregate fidelity and diversity. At a high level, we construct multiple datasets of similar fidelity but varying levels of diversity, and then verify that our metrics computed on these datasets agree with the expected trends. We also perform these experiments on different kinds of text data, demonstrating the general applicability of our method.

To that end, we begin by collecting data from the \textsc{M2D2 S2ORC} and \textsc{M2D2 Wikipedia} test splits in the Paloma benchmark \cite{magnusson2024palomabenchmarkevaluatinglanguage, reid2022m2d2}. \textsc{M2D2 S2ORC} contains Semantic Scholar papers and is organized by academic field, while \textsc{M2D2 Wikipedia} contains text from Wikipedia articles and is categorized according to the Wikipedia ontology. We construct a total of ten different collections of data: each of the first five collections contains ten randomly sampled categories from \textsc{M2D2 S2ORC}, while each of the remaining five consists of ten randomly sampled categories from \textsc{M2D2 Wikipedia}. All ten collections are then embedded using $\text{E5}_\text{small}$ \cite{wang2024textembeddingsweaklysupervisedcontrastive}, with mean-pooling over sentences to produce sentence embeddings of the data.

We now construct multiple datasets from each collection, controlling the diversity in each dataset by varying the number of categories included, and fixing fidelity across datasets by selecting an appropriate reference dataset for each collection. In particular, we would like the reference dataset to encompass all topics available in that collection, so that any other dataset composed of categories from that collection should have uniformly high fidelity with respect to the reference. A natural choice is therefore to pool together a subset of every category in the collection to serve as the reference. We collect only the first half of each category for the reference and construct all other datasets from the remaining samples, so as to avoid duplicate samples across evaluation and reference datasets. As a numerical baseline, we also compute our metrics between all remaining samples and the reference, since the remaining samples can be understood as being drawn from the same underlying distribution as the reference. These baseline values thus capture ideal fidelity and diversity for the given collection.

We then construct ten datasets of varying diversity from those remaining samples. In particular, for every $i \in [1, 10]$, we construct a dataset composed of the second half of samples in categories 1 through $i$, and refer to it as subset $i$. For instance, subset 1 contains exactly the second half of samples from category 1, while  subset 10 contains the second halves of all ten categories. We highlight that these subsets exhibit a nested structure, where subset $i+1$ contains all of subset $i$, and so on. By design, we expect the smaller subsets to be strictly less diverse than the larger ones, as they contain strictly fewer topics. However, all subsets should be of comparable fidelity with respect to the reference.

To apply our method to the constructed subsets, we first quantize each dataset using Gaussian mixture models (GMMs). Since the reference and subsets span a wide range of sizes, we independently fit a GMM for each and select the number of mixture components using the Bayesian Information Criterion (BIC). The fidelity and diversity scores are then computed directly on the mixing measure underlying the selected mixture model. For further details about this quantization step, see App.~\ref{app:m2d2-stats}.

\begin{figure}[h]
\centering
\begin{subfigure}{0.5\textwidth}
  \centering
  \includegraphics[width=.95\linewidth]{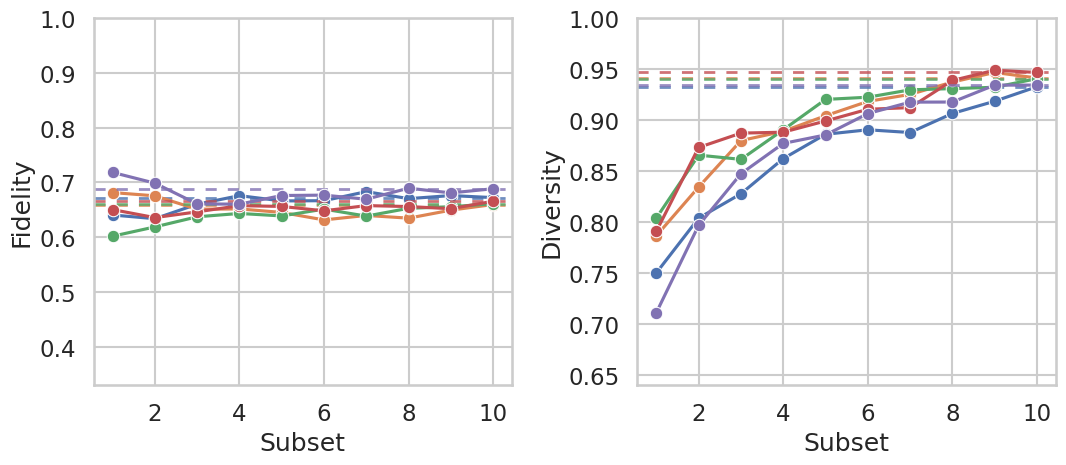}
  \caption{\textsc{M2D2 S2ORC}
  }

  \label{fig:s2}
\end{subfigure}%
\begin{subfigure}{0.5\textwidth}
  \centering
  \includegraphics[width=.95\linewidth]{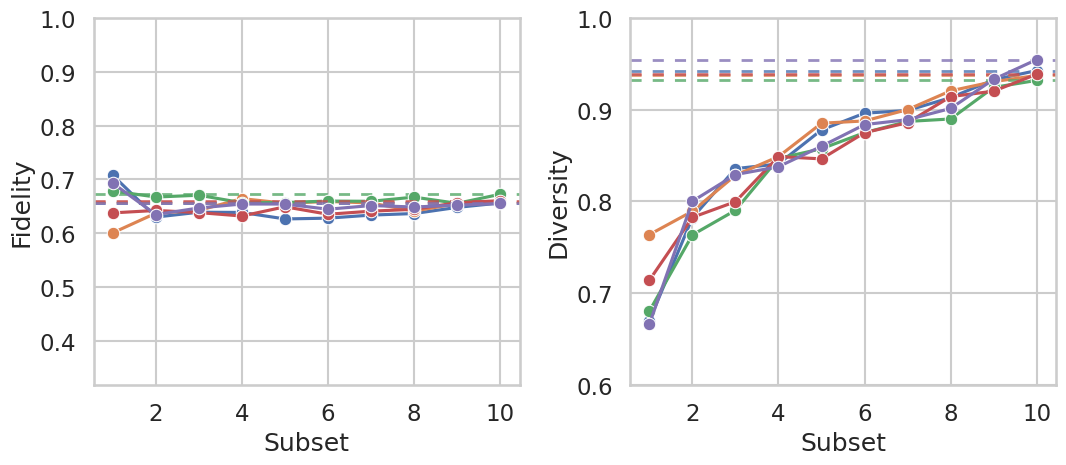}
  \caption{\textsc{M2D2 Wikipedia}
  }
  \label{fig:wiki}
\end{subfigure}
\caption{\small{Fidelity and diversity scores for collections \textcolor{RoyalBlue}{0}, \textcolor{BurntOrange}{1}, \textcolor{Green}{2}, \textcolor{BrickRed}{3}, and \textcolor{Purple}{4} of text from \textsc{M2D2 S2ORC} and \textsc{M2D2 Wikipedia}. The $x$-axes indicate the subset of the collection being measured against the reference. Dashed lines indicate baseline values of fidelity and diversity for each collection. See App.~\ref{app:m2d2_categories} for a detailed list of the ten categories in each collection, as well as bootstrapped confidence bands for all collections.}}
\label{fig:m2d2_metrics}
\end{figure}

Examining Fig.~\ref{fig:m2d2_metrics} and Table~\ref{tab:m2d2_correlations}, observe that the fidelity scores of subsets with respect to our chosen references do not typically exhibit significant association with the number of categories in the data, as expected. On the other hand, in all ten collections, the diversity score improves steadily and has a statistically significant OLS regression coefficient with respect to the number of categories. This is again consistent with our expectations and serves as evidence that \textbf{our metrics consistently disaggregate fidelity and diversity across different types of text data}. We also provide a more detailed discussion of these experiments with bootstrapped confidence intervals in App.~\ref{app:m2d2-stats}.

\subsubsection{Validating fidelity} 
\begin{figure}[h]
\centering
\begin{subfigure}{.5\textwidth}
  \centering
  \includegraphics[width=.95\linewidth]{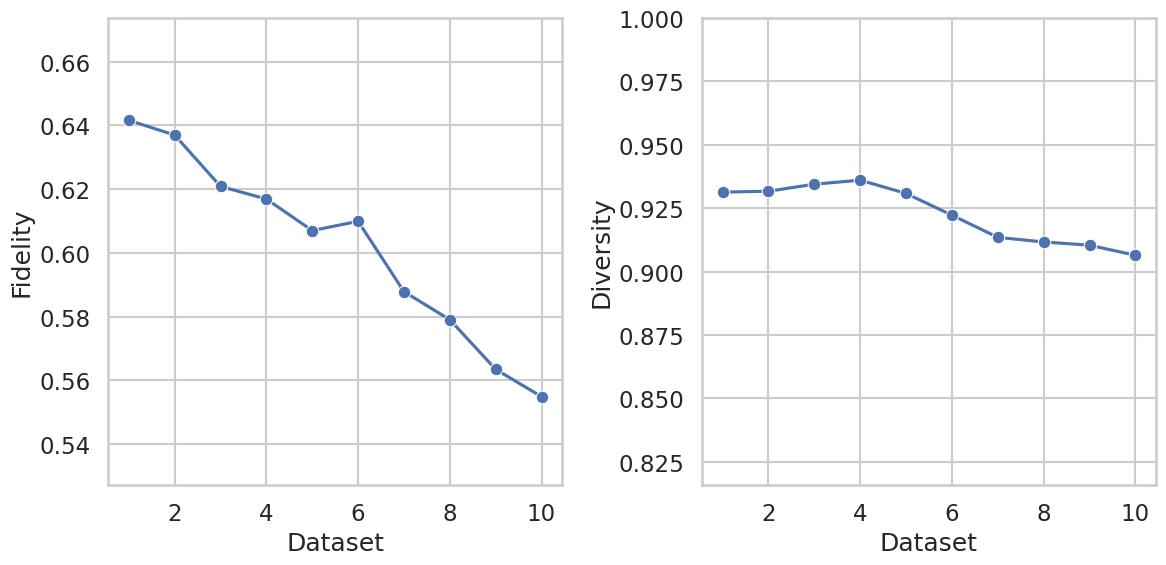}
  \caption{\textsc{M2D2 S2ORC}}
  \label{fig:s2_swap}
\end{subfigure}%
\begin{subfigure}{.5\textwidth}
  \centering
  \includegraphics[width=.95\linewidth]{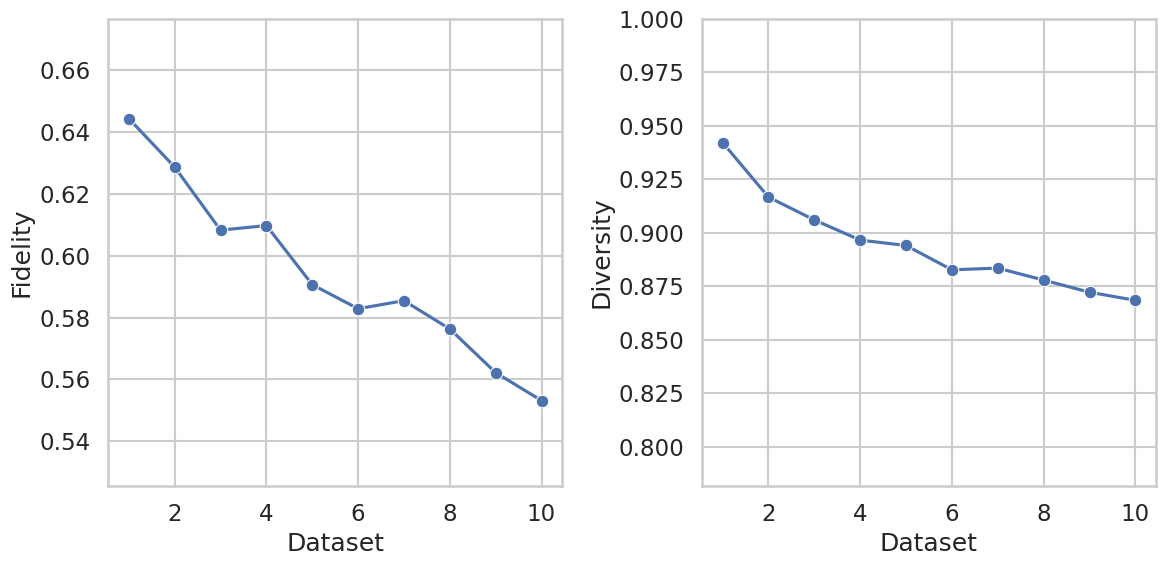}
  \caption{\textsc{M2D2 Wikipedia}}
  \label{fig:wiki_swap}
\end{subfigure}
\caption{\small{Fidelity \& diversity scores for controlled distribution-shift experiments. Fidelity scores for both collections exhibit significant negative association with the number of categories disjoint from the reference; computing the OLS regression coefficient $\beta$ and t-test $p$-value for the null hypothesis $\beta = 0$ gives $\beta=\mathbf{-0.0096}$ ($p<0.001$) for \textsc{M2D2 S2ORC} and $\beta=\mathbf{-0.0093}$ ($p<0.001$)} for \textsc{M2D2 Wikipedia}}
\label{fig:m2d2_swap}
\end{figure}

To verify whether our fidelity scores are actually sensitive to changes in fidelity, we also construct two experiments in which we incrementally shift the evaluation dataset away from the reference. More concretely, we select ten categories $C_1, \dots, C_{10}$ for the reference, as well as ten additional categories $D_1, \dots, D_{10}$ disjoint from the reference. Each dataset $i$ consists of data from subdomains $D_1, \dots D_i, C_{i+1}, \dots C_{10}$. Thus, lower values of $i$ correspond to datasets closer to the reference, while higher values correspond to being further from the reference. We again embed the data using $\text{E5}_\text{small}$ and fit spherical GMMs for all datasets. As shown in Fig.~\ref{fig:m2d2_swap}, our fidelity scores are capable of detecting gradual degradations in fidelity.

\begin{table}[t]
\centering
\small
\begin{subtable}[t]{0.48\linewidth}
\centering
\begin{tabular}{ccccc}
\toprule
& \multicolumn{2}{c}{Fidelity} & \multicolumn{2}{c}{Diversity} \\
\cmidrule(lr){2-3} \cmidrule(lr){4-5}
Coll. & $\beta$ & $p$ & $\beta$ & $p$ \\
\midrule
0 & +0.0011 & 0.586 & \textbf{+0.0203} & 0.001 \\
1 & -0.0031 & 0.083 & \textbf{+0.0157} & $<0.001$ \\
2 & \textbf{+0.0054} & 0.001 & \textbf{+0.0130} & $<0.001$ \\
3 & +0.0016 & 0.064 & \textbf{+0.0140} & $<0.001$ \\
4 & -0.0014 & 0.523 & \textbf{+0.0209} & 0.001 \\
\bottomrule
\end{tabular}
\caption{\textsc{M2D2 S2ORC}}
\end{subtable}
\hfill
\begin{subtable}[t]{0.48\linewidth}
\centering
\begin{tabular}{ccccc}
\toprule
& \multicolumn{2}{c}{Fidelity} & \multicolumn{2}{c}{Diversity} \\
\cmidrule(lr){2-3} \cmidrule(lr){4-5}
Coll. & $\beta$ & $p$ & $\beta$ & $p$ \\
\midrule
0 & -0.0021 & 0.456 & \textbf{+0.0246} & $<0.001$ \\
1 & +0.0037 & 0.064 & \textbf{+0.0191} & $<0.001$ \\
2 & -0.0009 & 0.329 & \textbf{+0.0244} & $<0.001$ \\
3 & \textbf{+0.0023} & 0.019 & \textbf{+0.0221} & $<0.001$ \\
4 & -0.0014 & 0.445 & \textbf{+0.0244} & $<0.001$ \\
\bottomrule
\end{tabular}
\caption{\textsc{M2D2 Wikipedia}}
\end{subtable}

\caption{Ordinary least squares (OLS) regression coefficients ($\beta$)  of fidelity-diversity metrics on subset index across collections. We also report t-test $p$-values, where the null hypothesis is $\beta = 0$. Bold entries indicate statistical significance ($p<0.05$).}
\label{tab:m2d2_correlations}
\end{table}

\subsection{Synthetic GSM8K experiments }\label{Syn}
A popular approach for training models given limited data is to augment available training data with \textit{synthetic data} from a strong existing model. Although many works explore the potential for significant boosts in performance from using high-quality synthetic data \cite{liu2023tinygsmachieving80gsm8k, yang2024syntheticcontinuedpretraining, su2025nemotroncctransformingcommoncrawl}, efficiently diagnosing synthetic data quality ahead of model training remains a key practical challenge. Our metric enables direct quantitative assessments of diversity in synthetic data, which we find to be correlated with the  performances of models trained on that data.

\subsubsection{Setup}

We generate synthetic GSM8K-style datasets of varying diversities using a combination of GPT-4.1, -5.2, and -5.4 \cite{openai2024gpt4technicalreport, singh2026openaigpt5card}. To generate synthetic problems, we provide GPT-4.1 with a problem from GSM8K, the ``seed'' problem, and prompt it to generate a structurally similar rewrite of the seed problem. Using $k$ seed problems from the original $n$ problems in GSM8K, we generate a synthetic dataset by prompting GPT-4.1 $(n-k)/k$ times for each seed problem, generating multiple synthetic examples based on each seed. By varying the number of seed problems $k$, we  control the amount of diversity in the resulting synthetic dataset: if $k=1$, then every synthetic problem is a variant of just one problem from the original GSM8K; if $k=n$, the resulting dataset should have comparable diversity to the original GSM8K. We generate synthetic datasets for each $k \in \{0.1\%, 0.2\%, 0.5\%, 1\%, 2\%, 5\%, 10\%\} \times n$, and refer to each synthetic dataset by the percentage of GSM8K training samples in its seed data. For each generated problem, we then generate step-by-step solutions using GPT-5, adopting a similar prompting strategy as \cite{toshniwal2024openmathinstruct2acceleratingaimath}. For more details about the synthetic data generation pipeline, see App.~\ref{app:synth_data}. 


Applying our metric requires an appropriate reference. The GSM8K training split is inadequate as a reference for comparing synthetic datasets, since it confounds the distributional shift in linguistic style arising from the synthetic data generation process with differences in underlying structure of the math problems that we wish to measure. Instead, we generate an additional fully synthetic reference dataset using the entire GSM8K training split as the seed ($k=n$): for each GSM8K training example, we generate a single synthetic problem-solution pair. Critically, we only include the synthetic generations in this reference dataset. Using a fully synthetic reference allows us to better isolate differences in structural diversity when comparing to other synthetic datasets. 

\begin{figure}[h]
\centering
\begin{subfigure}{.66\textwidth}
  \centering
  \includegraphics[width=.9\linewidth]{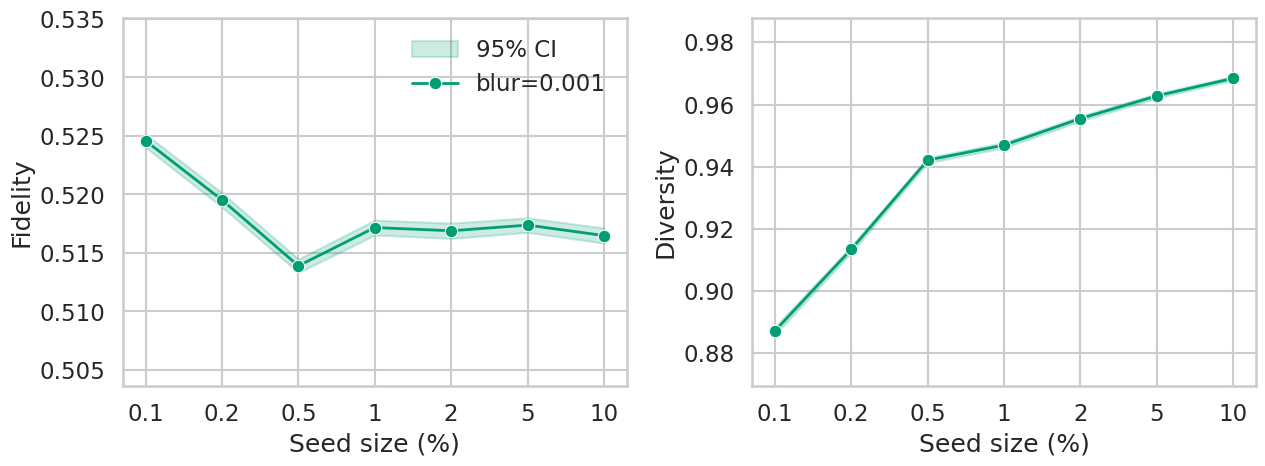}
  \caption{Fidelity + diversity of synthetic GSM8K-style datasets}
  \label{fig:gsm8k_metrics}
\end{subfigure}%
\begin{subfigure}{.33\textwidth}
  \centering
  \includegraphics[width=.9\linewidth]{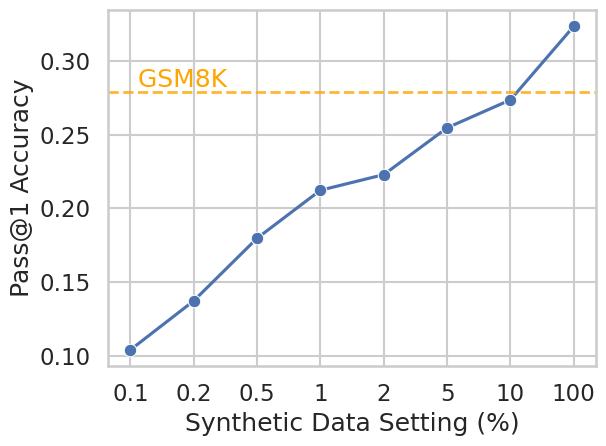}
  \caption{Llama-3.2-1B}
  \label{fig:gsm8k_test_acc}
\end{subfigure}
\caption{ \small{The $x$-coordinates correspond to the synthetic dataset being evaluated or finetuned on, identified by the percentage of GSM8K training samples present in the seed data. \textbf{(a)} Fidelity-diversity scores with OT approximated by debiased Sinkhorn divergences, with entropic regularization hyperparameter $=0.001$. The shaded regions in both plots correspond to 95\% confidence intervals computed from bootstrapped estimates of sample variance. Diversity scores exhibit strong correlation with downstream test accuracies (Pearson $r = \mathbf{+0.973}$ and $p$-value $< 0.001$); fidelity scores do not exhibit statistically-significant correlation ($r = -0.670$, $p = 0.099$). \textbf{ (b) }Exact-match accuracies of Llama-3.2-1B base models on the GSM8K test split after finetuning on synthetic datasets. The dashed orange line indicates test accuracy achieved by a base model finetuned on the GSM8K training split.}}
\label{fig:gsm8k}
\end{figure}

We then embed all datasets using $\text{E5}_\text{small}$ with mean-pooling over problems and solutions to produce one embedding for each problem and each solution. We omit the quantization step, as we found that the synthetic data embeddings are not amenable to common clustering methods like $k$-means, GMMs, or HDBSCAN, suggesting that the embeddings themselves do not contain well-separated clusters (for further discussion of this point, see App.~\ref{app:gsm8k_clustering}). We then compute fidelity-diversity scores on the embeddings themselves, with uniform weights and Euclidean distances between embeddings serving as the ground metric. While we can efficiently compute exact fidelity scores, we approximate the diversity scores via debiased Sinkhorn divergences \cite{feydy2019interpolating} 
(Fig.~\ref{fig:gsm8k_metrics}).

To investigate the relationship between training data diversity and downstream performance, we perform supervised finetuning (SFT) on a Llama-3.2-1B base model \cite{llama} with synthetic data. See Fig.~\ref{fig:gsm8k_test_acc} for model performance on the GSM8K test set after finetuning on different synthetic datasets, including the reference dataset seeded with the full GSM8K training split (labeled as 100\%). Finally, for comparison, we also finetune a base model on the original GSM8K training split and evaluate it on the test split. We provide further details on finetuning in App.~\ref{app:sft}.

\subsubsection{Discussion}

We find that our metrics effectively detect differences in underlying structural diversity across synthetic datasets, yielding measurements which correlate strongly with downstream model performance.

Observe that the base model finetuned on the fully-synthetic reference dataset (100\%) actually achieves stronger GSM8K performance than the one finetuned on the original GSM8K training set. We reproduce this finding on a larger Llama-3.2-3B model (Fig.~\ref{fig:gsm8k_llama_test-acc}), suggesting that our synthetic data generation process does in fact produce high-quality data given sufficiently diverse seed data. This points to improvements in model performance being the result of increasing data diversity, rather than merely the inclusion of more GSM8K training examples. 

Meanwhile, diversity scores across the seven datasets exhibit strong correlation with downstream test accuracies of models finetuned on the data ($r  = \mathbf{+0.973}$ and $p$-value $< 0.001$). In contrast, fidelity scores do not exhibit statistically-significant correlation ($r  = -0.670$, $p = 0.099$), further supporting the premise that our metrics are capable of disaggregating fidelity and diversity. We additionally verify that the fidelity metrics are responsive to failures of fidelity in this setting (App.~\ref{app:gsm8k_fidelity}).

The sensitivity of our metrics on these synthetic math datasets is particularly notable, considering that the embeddings appear to represent information about problem structure in a manner so subtle that conventional clustering methods fail.

\section{Conclusion}  
The aim of our work
has been to develop fidelity-diversity metrics
for comparing text datasets. On the one hand, we introduce
new constructions for this problem by leveraging
  optimal transport   between mixing measures. On the other hand, 
we show that modern embedding models are sufficiently
expressive for our metrics to detect human-interpretable 
variations in fidelity and diversity within large-scale datasets. 
While we expect our metrics to be applicable for a variety of tasks such as
language model evaluation, our primary experiments
focused on leveraging them to quantify the quality of synthetically-generated datasets.


Finally, we note that there is a growing
literature emphasizing the fragility of 
evaluation metrics for text data arising from 
open-ended language models~\citep{stein2023exposing, raisa2025position}. 
Though we do not propose solutions to these challenges, we 
hope our work nevertheless demonstrates the utility of such
metrics in current generative modeling applications.
 
\section*{Acknowledgements}
A. Wang gratefully acknowledges the support of the Juris Hartmanis/John Hopcroft fellowship. T. Manole gratefully acknowledges the support of a Norbert Wiener postdoctoral fellowship. F. Bunea’s research has been supported in part by NSF-DMS 2515156.

\bibliographystyle{abbrv}
\bibliography{ref}

\appendix
 
\section{Comparisons to Precision \& Recall} \label{app:pr}
While our metrics bear a strong resemblance to the precision and recall metrics put forth in \cite{le2024exploring}, we point out conceptual differences in our approaches. At a high level, their precision and recall scores quantify overlap in the supports of the evaluation and reference distributions $P$ and $Q$. Precision captures the proportion of points drawn from $P$ which lie in the support of $Q$, while recall captures the mass under $Q$ which is covered by the support of $P$:
\begin{equation}
\label{eq:pr}
\begin{aligned}
&\text{Precision} = \Pr_{u \sim P}[u \in \supp Q] = P(\supp Q)\\
&\text{Recall} = \Pr_{v \sim Q}[v \in \supp P] = Q(\supp P)
\end{aligned}
\end{equation}
In contrast, our fidelity score quantifies how far points drawn from $P$ lie from the support of $Q$, while diversity quantifies mass discrepancies between $P$ and $Q$. In particular, our diversity score is designed to capture differences in mass distribution, even over the same support, whereas the original definitions of precision and recall~\eqref{eq:pr} are entirely insensitive to mass discrepancies. Despite this, we find that the $k$-nearest-neighbor support estimation procedure proposed in \cite{le2024exploring} makes their precision-recall metrics sensitive to mass discrepancies in practice.

\begin{figure*}[h]
    \centering
    \begin{adjustbox}{max width=\textwidth}
    \begin{tblr}{ Q[2.5cm,valign=m,halign=c] Q[10cm,valign=h,halign=c] Q[10cm,valign=h,halign=c] }
        & \textbf{\Large $\text{E5}_{\text{small}}$} & \textbf{\Large GPT-2} \\
        \cmidrule[lr]{2-2}
        \cmidrule[lr]{3-3}
        \textbf{\Large (Fidelity, Diversity)}
    
        &
        \begin{subfigure}{.48\linewidth}
          \centering
          \vspace{2em}
          \includegraphics[width=\linewidth]{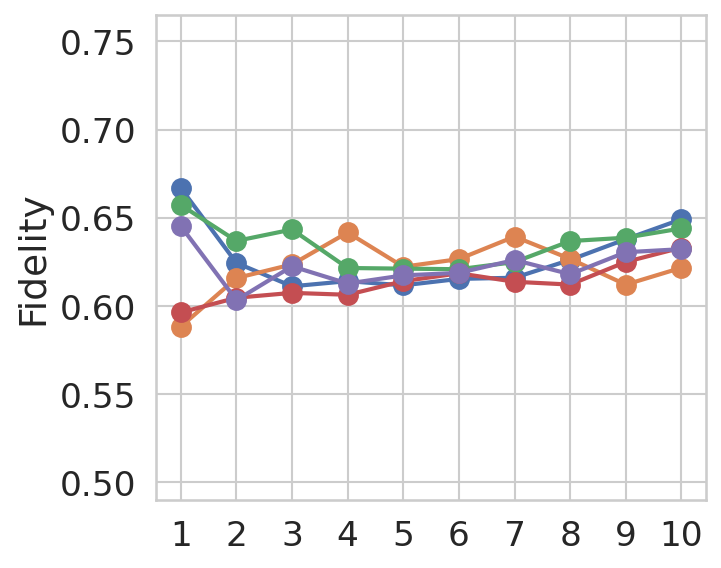}
        \end{subfigure}%
        \hfill
        \begin{subfigure}{.48\linewidth}
          \centering
          \includegraphics[width=\linewidth]{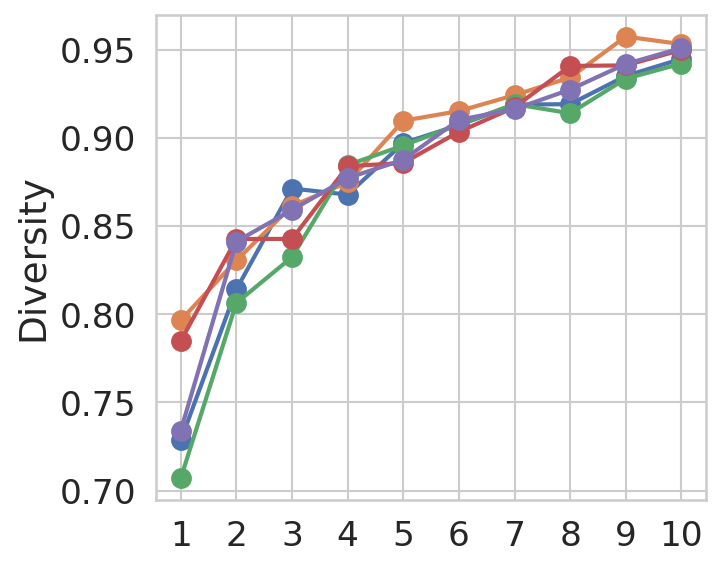}
        \end{subfigure}
        &
        \begin{subfigure}{.48\linewidth}
          \centering
          \vspace{2em}
          \includegraphics[width=\linewidth]{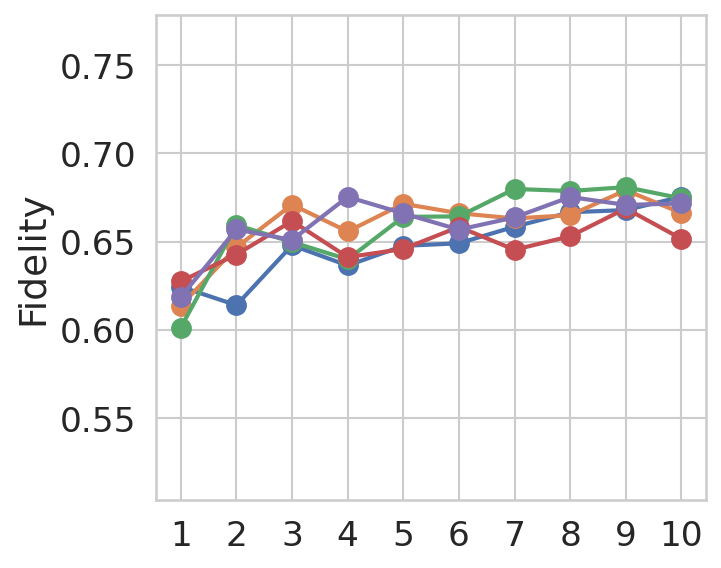}
        \end{subfigure}%
        \hfill
        \begin{subfigure}{.48\linewidth}
          \centering
          \includegraphics[width=\linewidth]{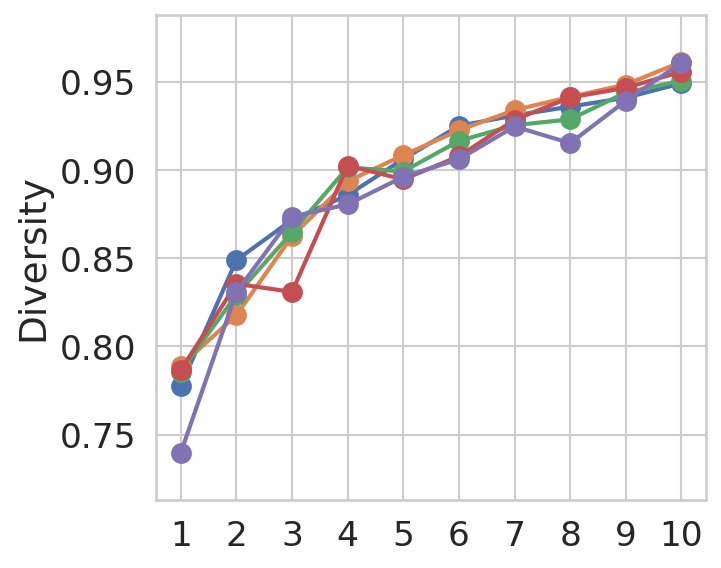}
        \end{subfigure}
        \\
        \textbf{\Large (Precision, Recall)}
        &
        \begin{subfigure}{.48\linewidth}
          \centering
          \vspace{2em}
          \includegraphics[width=\linewidth]{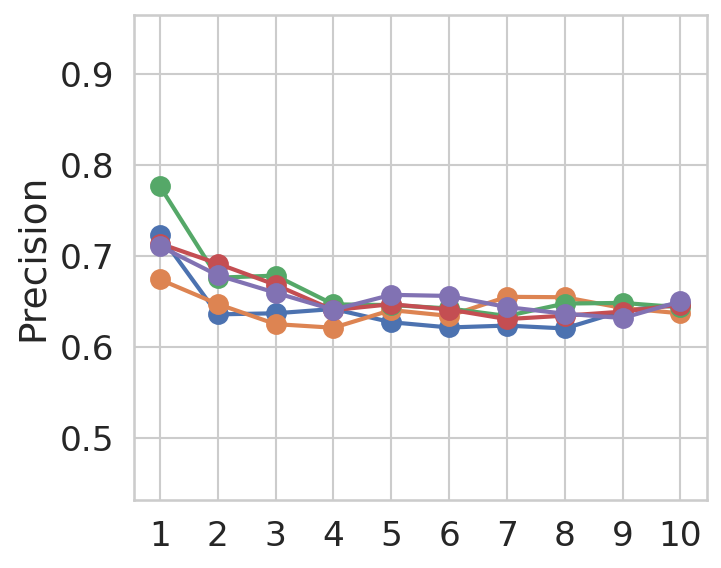}
        \end{subfigure}%
        \hfill
        \begin{subfigure}{.48\linewidth}
          \centering
          \includegraphics[width=\linewidth]{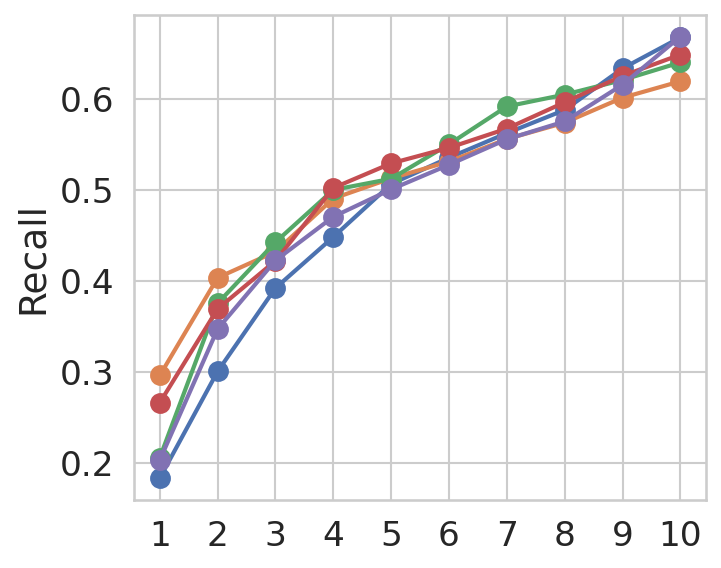}
        \end{subfigure}
        &
        \begin{subfigure}{.48\linewidth}
          \centering
          \vspace{2em}
          \includegraphics[width=\linewidth]{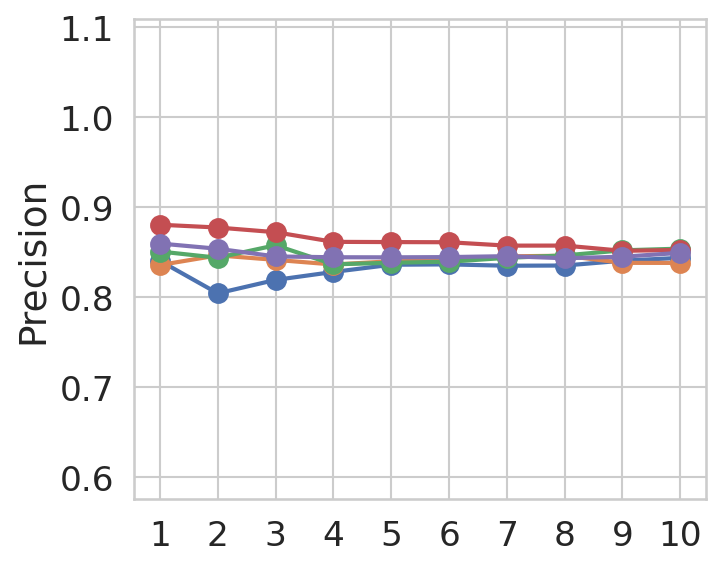}
        \end{subfigure}%
        \hfill
        \begin{subfigure}{.48\linewidth}
          \centering
          \includegraphics[width=\linewidth]{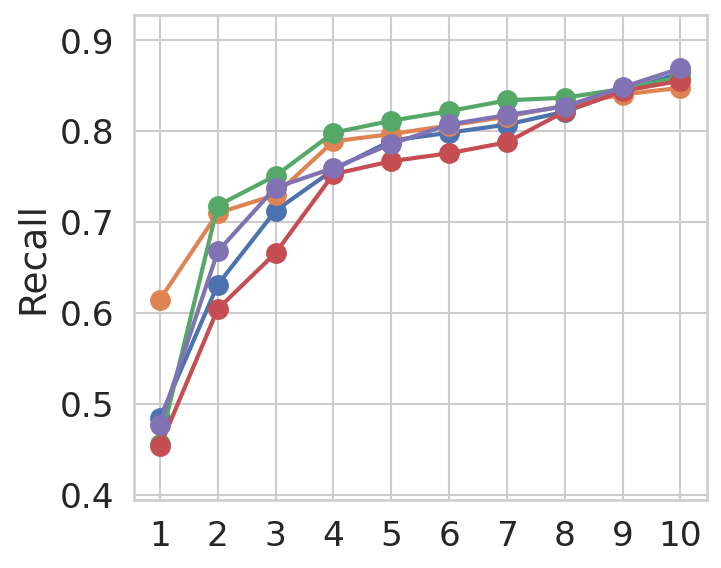}
        \end{subfigure}
    \end{tblr}
    \end{adjustbox}

    \caption{
        Comparison of our metric (top) with precision and recall (bottom) using $\text{E5}_{\text{small}}$ (left) and GPT-2 (right) embeddings.
        Each cell shows the corresponding metrics computed on collections \textcolor{RoyalBlue}{0}, \textcolor{BurntOrange}{1}, \textcolor{Green}{2}, \textcolor{BrickRed}{3}, and \textcolor{Purple}{4} of text from \textsc{M2D2 Wikipedia}. The $x$-axes indicate the subset of the collection being measured against the reference.
    }
    \label{fig:metric_pr_comparison}
\end{figure*}

We empirically compare both sets of metrics across several datasets, as well as across $\text{E5}_{\text{small}}$ and GPT-2 text embeddings (Figure~\ref{fig:metric_pr_comparison}). In practice, both our metrics and theirs yield similar results on the five collections of text from \textsc{M2D2 Wikipedia} used in Section~\ref{M2D2}. We use the same partitioning of each collection into reference and evaluation subsets as in Section~\ref{M2D2}. Both methods also perform similarly when computed on different text embeddings for typical text datasets. 

\section{M2D2 experiments}

\subsection{Data} \label{app:m2d2_categories}

\textsc{M2D2 S2ORC} consists of text sampled from academic papers organized into a two-layer categorization by academic field. Top-level domains were first provided in \textsc{S2ORC} \cite{lo2020s2orcsemanticscholaropen}, while subdomains follow the paper's Arxiv categorization. \textsc{M2D2 Wikipedia} contains text from Wikipedia articles categorized by the top two levels of the Wikipedia ontology.

Each of the following collections consist of ten subdomains sampled uniformly at random (without replacement) from all available subdomains in \textsc{M2D2 S2ORC} or \textsc{M2D2 Wikipedia}.

\subsubsection{\textsc{M2D2 S2ORC} collections}
\paragraph{Collection 0:} math.PR, cs.PL, q-fin.CP, cond-mat.quant-gas, eess.SP, stat.ML, q-fin.PR, q-bio.NC, nucl-th, physics.atm-clus
\paragraph{Collection 1:} physics.optics, math.RA, physics.class-ph, econ.TH, q-bio.MN, hep-th, Philosophy, physics\_l1, nucl-ex, math.NA
\paragraph{Collection 2:} math.CO, cond-mat\_l1, math.RA, stat.ML, cs.CE, plasm-ph, cs.GR, atom-ph, math.CA, math.SP
\paragraph{Collection 3:}cs.CV, cs.NA, q-fin.PM, stat.ML, physics.app-ph, q-bio.CB, eess.AS, math.FA, math.QA, math.CA
\paragraph{Collection 4:} math.CA, cond-mat.str-el, Art, astro-ph.IM, q-fin.CP, math\_l1, math.SP, astro-ph.EP, q-fin.GN, math.OC

For the fidelity validation experiment, the ten \textsc{M2D2 S2ORC} reference subdomains were all selected from the top-level 
computer science domain (\textit{cs.AI, cs.AR.str-el, cs.CC, cs.CE, cs.CG, cs.CL, cs.CR, cs.CV, cs.CY, and cs.DB}). The ten disjoint subdomains were selected to minimize topical overlap with the reference, and included subdomains from physics, philosophy, and art (\textit{astro-ph.CO, atom-ph, chen-ph, Philosophy, cond-mat.soft, astro-ph.EP, Art, cond-mat.quant-gas, astro-ph.HE, and cond-mat.stat-mech}).

\subsubsection{\textsc{M2D2 Wikipedia} collections}
\paragraph{Collection 0:} Culture\_and\_the\_arts\_\_Games\_and\_Toys, Health\_and\_fitness\_\_Exercise, History\_and\_events\_\_By\_continent, Natural\_and\_physical\_sciences\_\_Nature, Religion\_and\_belief\_systems\_\_Belief\_systems, Culture\_and\_the\_arts\_\_Culture\_and\_Humanities, Human\_activites\_\_Human\_activities, Society\_and\_social\_sciences\_\_Society, Culture\_and\_the\_arts\_\_Mass\_media, Culture\_and\_the\_arts\_\_The\_arts\_and\_Entertainment

\paragraph{Collection 1:}Natural\_and\_physical\_sciences\_\_Biology, Health\_and\_fitness\_\_Health\_science, Culture\_and\_the\_arts\_\_Culture\_and\_Humanities, Health\_and\_fitness\_\_Self\_care, Religion\_and\_belief\_systems\_\_Major\_beliefs\_of\_the\_world, Health\_and\_fitness\_\_Human\_medicine, Religion\_and\_belief\_systems\_\_Allah, Mathematics\_and\_logic\_\_Logic, Health\_and\_fitness\_\_Exercise, Technology\_and\_applied\_sciences\_\_Engineering
\paragraph{Collection 2:}Culture\_and\_the\_arts\_\_Performing\_arts, Health\_and\_fitness\_\_Self\_care, Natural\_and\_physical\_sciences\_\_Physical\_sciences, Health\_and\_fitness\_\_Exercise, Human\_activites\_\_Human\_activities, Philosophy\_and\_thinking\_\_Thinking, Society\_and\_social\_sciences\_\_Social\_sciences, Religion\_and\_belief\_systems\_\_Major\_beliefs\_of\_the\_world, Culture\_and\_the\_arts\_\_Sports\_and\_Recreation, Health\_and\_fitness\_\_Health\_science
\paragraph{Collection 3:} Religion\_and\_belief\_systems\_\_Major\_beliefs\_of\_the\_world, Natural\_and\_physical\_sciences\_\_Physical\_sciences, Culture\_and\_the\_arts\_\_Sports\_and\_Recreation, Technology\_and\_applied\_sciences\_\_Engineering, History\_and\_events\_\_By\_region, History\_and\_events\_\_By\_continent, History\_and\_events\_\_By\_period, Technology\_and\_applied\_sciences\_\_Computing, Natural\_and\_physical\_sciences\_\_Biology, Culture\_and\_the\_arts\_\_Games\_and\_Toys
\paragraph{Collection 4:} Culture\_and\_the\_arts\_\_The\_arts\_and\_Entertainment, History\_and\_events\_\_By\_continent, Health\_and\_fitness\_\_Self\_care, Religion\_and\_belief\_systems\_\_Belief\_systems, Culture\_and\_the\_arts\_\_Games\_and\_Toys, Mathematics\_and\_logic\_\_Logic, History\_and\_events\_\_By\_period, Culture\_and\_the\_arts\_\_Culture\_and\_Humanities, Culture\_and\_the\_arts\_\_Sports\_and\_Recreation, Culture\_and\_the\_arts\_\_Mass\_media

For the fidelity validation experiment, the ten \textsc{M2D2 Wikipedia} reference subdomains were selected from the natural and applied sciences (\textit{Biology, Earth_sciences, Nature, Physical_sciences, Mathematics, Logic, Agriculture, Computing, Engineering, Transport}), while the disjoint subdomains were focused on the arts, history, and culture (\textit{Culture_and_Humanities, Games_and_Toys, Mass_media, Performing_arts, Culture_and_the_arts__Sports_and_Recreation, The_arts_and_Entertainment, Visual_arts, History_and_events_By_continent, History_and_events__By_period, History_and_events_By_region}).

\subsection{Statistical methods} \label{app:m2d2-stats}
All \textsc{M2D2} datasets were quantized using Gaussian Mixture Models (GMMs) with spherical covariances. We fit the GMMs using maximum likelihood
estimation, approximated via the Expectation-Maximization algorithm. 
The number of mixture components $k$ was independently selected for each dataset by minimizing the Bayesian Information Criteria (BIC) across models fitted with $$k \in \{5, 10, 15, 20, 25, 30, 40, 50, 60, 70, 80, 90, 100\}.$$ 

\begin{figure}[h]
\centering
\begin{subfigure}{\textwidth}
  \centering
  \includegraphics[width=\linewidth]{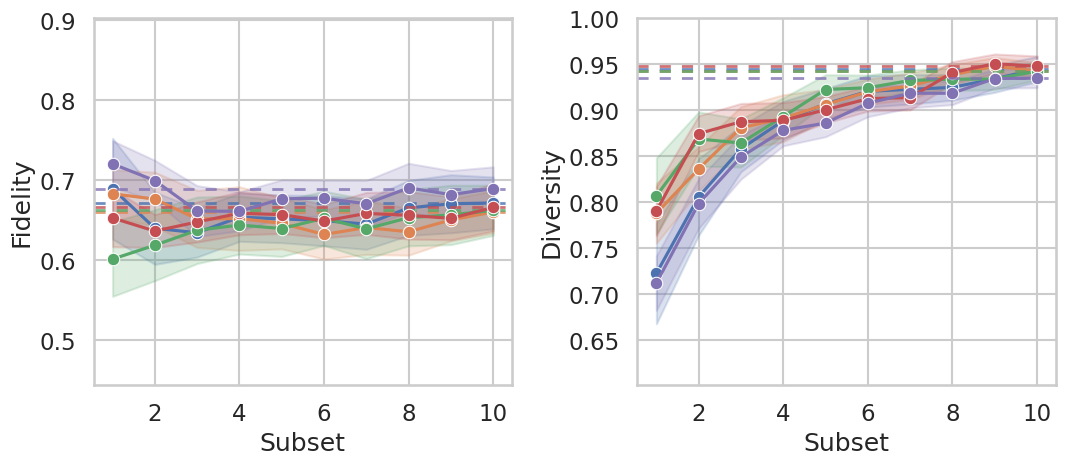}
  \caption{\textsc{M2D2 S2ORC}}
  \label{fig:s2_cis}
\end{subfigure}%
\\[2ex]
\begin{subfigure}{\textwidth}
  \centering
  \includegraphics[width=\linewidth]{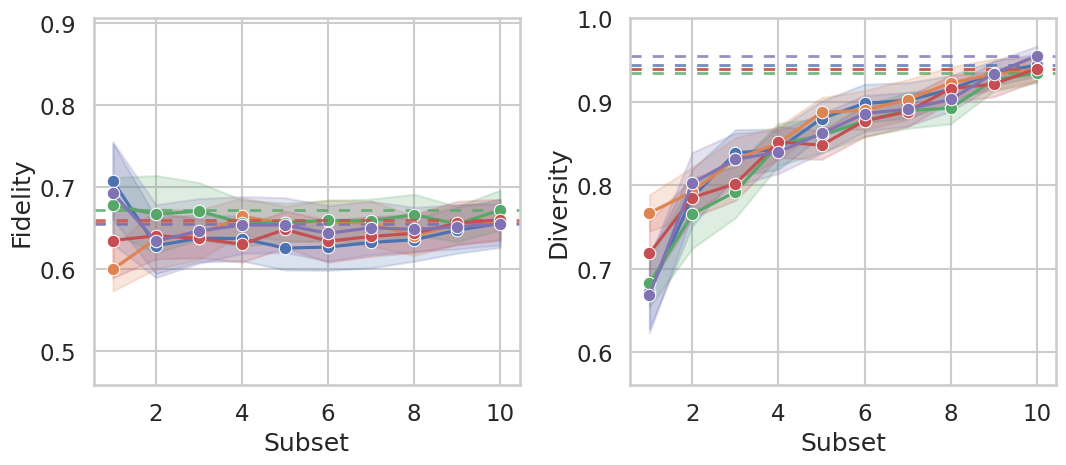}
  \caption{\textsc{M2D2 Wikipedia}}
  \label{fig:wiki_cis}
\end{subfigure}
\caption{Bootstrapped estimates of 95\% confidence intervals of fidelity and diversity scores, indicated by the shaded regions. The solid and dashed lines are identical to those in Fig.~\ref{fig:m2d2_metrics}}.
\label{fig:m2d2_cis}
\end{figure}

We adopt a ground metric  for atoms $\theta_1 = (\mu_1, \sigma_1^2 \mathrm{Id}_z)$, $\theta_2 = (\mu_2, \sigma_2^2 \mathrm{Id}_z)$ which was previously
 proposed in the theoretical literature
on location-scale Gaussian mixture models~\citep{hardt2015tight,ho2016convergence,manole2020uniform}:
\[
d(\theta_1, \theta_2) = \sqrt{\|\mu_1 - \mu_2\|_2^2 + z|\sigma_1^2 - \sigma_2^2|}.
\]  
We report estimates of our metrics
together with 95\% Gaussian variability
intervals of the form $\hat\theta \pm 1.96 \hat\sigma$, 
where $\hat\theta$ denotes a fidelity or diversity metric, 
and $\hat\sigma^2$ denotes a nonparametric bootstrap estimate
of its variance~\citep{wasserman2004} (Fig.~\ref{fig:m2d2_cis}). 
We use 500 bootstrap replications for each variance estimate. 



\section{Synthetic GSM8K experiments}

\subsection{Synthetic GSM8K data generation} \label{app:synth_data}
GSM8K \cite{cobbe2021training} is a curated collection of 8.8k grade-school math word problems created by human problem writers, divided into a 7.4k training split and a 1.3k test split. Each example consists of a problem-solution pair, where the problem can be solved using some sequence of the four elementary operations (addition, subtraction, multiplication, division) and has a well-defined positive integer answer. The solutions contain the key reasoning steps expressed in natural language, followed by the final answer.

To elaborate on the synthetic dataset construction, for each of the 
\{0.1\%, 0.2\%, 0.5\%, 1\%, 2\%, 5\%, 10\%\} synthetic datasets, we used the first \{0.1\%, 0.2\%, 0.5\%, 1\%, 2\%, 5\%, 10\%\} problems in the GSM8K training set as the seed. Within each dataset, we generated the same number of synthetic rewrites for every seed problem; to ensure that the final datasets were all of similar sizes, this meant generating \{999, 499, 199, 99, 49, 19, 9\} synthetic rewrites for each seed problem. Then, for each dataset, we appended the seed problems to the generated synthetic problems and shuffled the problems before storing them. 

The synthetic data generation pipeline consisted of three steps. We first generated questions using GPT-4.1 with temperature=1.5 to encourage variety in the generations. We then generated solutions for each synthetic question  using GPT-5.2 with reasoning effort=``high'' and verbosity=``medium'' to encourage correct solutions with clear reasoning steps. Finally, we checked for proper answer formatting and whether the final answer was a positive integer. Occasionally, a generated question would yield a nonpositive or noninteger answer. For all invalid problem-solution pairs, we used GPT-5.4 (again with reasoning effort=``high'' and verbosity=``medium'') to adjust the numbers in the problem and solution so as to yield positive integer answers, repeating this procedure as needed. See the following subsections for the exact prompt templates used in each step.

\subsubsection{Question generation prompt}
\begin{lstlisting}[frame=single]
Help the user create a new math dataset by generating problems one at a time. You will be given a reference problem from an existing dataset, from which you should generate a similar problem. 
Be creative with the names and contexts. Each problem you generate should be different. The problem must have a positive integer answer.

Here are some examples of how to complete this task.

Problem: 
Natalia sold clips to 48 of her friends in April, and then she sold half as many clips in May. How many clips did Natalia sell altogether in April and May?

Write another problem similar to this one.
Ryan sold 8 packs of gum in September, and then he sold twice as many packs in October. How many packs did Ryan sell in total during September and October?

Problem: 
Weng earns $12 an hour for babysitting. Yesterday, she just did 50 minutes of babysitting. How much did she earn?

Write another problem similar to this one.
Ali makes $48 an hour as a tutor. On Mondays, she tutors for 45 minutes. How much does she make on Mondays?

Problem: 
Betty is saving money for a new wallet which costs $100. Betty has only half of the money she needs. Her parents decided to give her $15 for that purpose, and her grandparents twice as much as her parents. How much more money does Betty need to buy the wallet?

Write another problem similar to this one.
Andre works at a bike repair shop with Tom and Brady. On Tuesday, there are twenty bikes they need to repair. By noon, Andre has repaired a quarter of them. Tom has repaired three, and Brady has repaired double that. How many remaining bikes do they need to repair in the afternoon?

Here is the problem from the user: 
{question}

Write another problem similar to this one. Start directly with the problem statement and DO NOT include any phrases such as "Here is a new problem similar to a given one".
After the problem is generated, finish your response right away.
\end{lstlisting}

\subsubsection{Solution generation prompt}
\begin{lstlisting}
Here are some examples of problems and solutions, followed by a new problem that you need to solve.
Make sure to include each step in your reasoning. You may use either x or * to indicate multiplication, and / for division. 
Write your answer (and only the numeric answer) on a new line beginning with "#### ".


Problem: 
Natalia sold clips to 48 of her friends in April, and then she sold half as many clips in May. How many clips did Natalia sell altogether in April and May?

Solution:
Natalia sold 48/2 = 24 clips in May.
Natalia sold 48+24 = 72 clips altogether in April and May.
#### 72


Problem:
Weng earns $12 an hour for babysitting. Yesterday, she just did 50 minutes of babysitting. How much did she earn?

Solution:
Weng earns 12/60 = $0.2 per minute.
Working 50 minutes, she earned 0.2 x 50 = $10.
#### 10


Problem:
Betty is saving money for a new wallet which costs $100. Betty has only half of the money she needs. Her parents decided to give her $15 for that purpose, and her grandparents twice as much as her parents. How much more money does Betty need to buy the wallet?

Solution:
In the beginning, Betty has only 100 / 2 = $50.
Betty's grandparents gave her 15 * 2 = $30.
This means Betty needs 100 - 50 - 30 - 15 = $5 more.
#### 5


Problem:
{synthetic question}

Solution:
\end{lstlisting}

\subsubsection{Problem adjustment prompt}
\begin{lstlisting}
Help the user fix a math problem to have positive integer solutions by changing the numbers involved.
Make sure to leave the rest of the problem and solution unchanged. Here are some examples of how to complete this task.


Problem: Six coworkers go out for lunch and order the following items: 6 slices of pizza at $2 each; 3 bowls of salad at $3.50 each; 6 bottles of juice at $1.25 each; and 2 plates of pasta at $4.75 each. If they split the total bill evenly, how much does each coworker need to pay?

Solution: Pizza costs 6 x $2 = $12.  
Salad costs 3 x $3.50 = $10.50.  
Juice costs 6 x $1.25 = $7.50.  
Pasta costs 2 x $4.75 = $9.50.  

Total bill is $12 + $10.50 + $7.50 + $9.50 = $39.50.  
Split among 6 coworkers: $39.50 / 6 = $6.5833..., which rounds to $6.58.

#### 6.58

Fixed Problem: Six coworkers go out for lunch and order the following items: 6 slices of pizza at $2 each; 3 bowls of salad at $3.50 each; 6 bottles of juice at $0.75 each; and 2 plates of pasta at $4.50 each. If they split the total bill evenly, how much does each coworker need to pay?

Fixed Solution: Pizza costs 6 x $2 = $12.  
Salad costs 3 x $3.50 = $10.50.  
Juice costs 6 x $0.75 = $4.50.  
Pasta costs 2 x $4.50 = $9.  

Total bill is $12 + $10.50 + $4.50 + $9 = $36.  
Split among 6 coworkers: $36 / 6 = $6.

#### 6


Problem: In a zoo, 16 parrots are sharing 15 bowls of seeds. 7 bowls each have 13 scoops, where each scoop contains 11 seeds. The other bowls each contain 8 scoops, and each scoop has 15 seeds. If the total seeds are divided equally among all the parrots, how many seeds does each parrot get?

Solution: 7 bowls have 13 * 11 = 143 seeds each, so they have 7 * 143 = 1001 seeds total.  
The other 15-7=8 bowls have 8 * 15 = 120 seeds each, so they have 8 * 120 = 960 seeds total.  
Altogether there are 1001 + 960 = 1961 seeds.  
Dividing equally among 16 parrots gives 1961 / 16 = 122.5625 seeds per parrot.

#### 122.5625


Fixed Problem: In a zoo, 16 parrots are sharing 15 bowls of seeds. 7 bowls each have 9 scoops, where each scoop contains 16 seeds. The other bowls each contain 8 scoops, and each scoop has 15 seeds. If the total seeds are divided equally among all the parrots, how many seeds does each parrot get?

Fixed Solution: 7 bowls have 9 * 16 = 144 seeds each, so they have 7 * 144 = 1008 seeds total.  
The other 15-7=8 bowls have 8 * 15 = 120 seeds each, so they have 8 * 120 = 960 seeds total.  
Altogether there are 1008 + 960 = 1968 seeds.  
Dividing equally among 16 parrots gives 1968 / 16 = 123 seeds per parrot.

#### 123


Here is the problem and solution from the user:

Problem: {invalid synthetic question}

Solution: {previously generated solution}


\end{lstlisting}

\subsection{Clustering} \label{app:gsm8k_clustering}
Fitting a GMM with BIC on the embedded synthetic datasets typically resulted in a number of mixture components on the order of the embedding dimension. Further exploratory data analysis suggested that the data lacks discernable clusters when mapped into the embedding space. This motivated our decision to compute our metrics directly on the empirical measure, as opposed to atoms derived from clustering, as in the previous experiments.


\subsection{Validating fidelity} \label{app:gsm8k_fidelity}
\begin{figure}[H]
\centering
  \includegraphics[width=.8\linewidth]{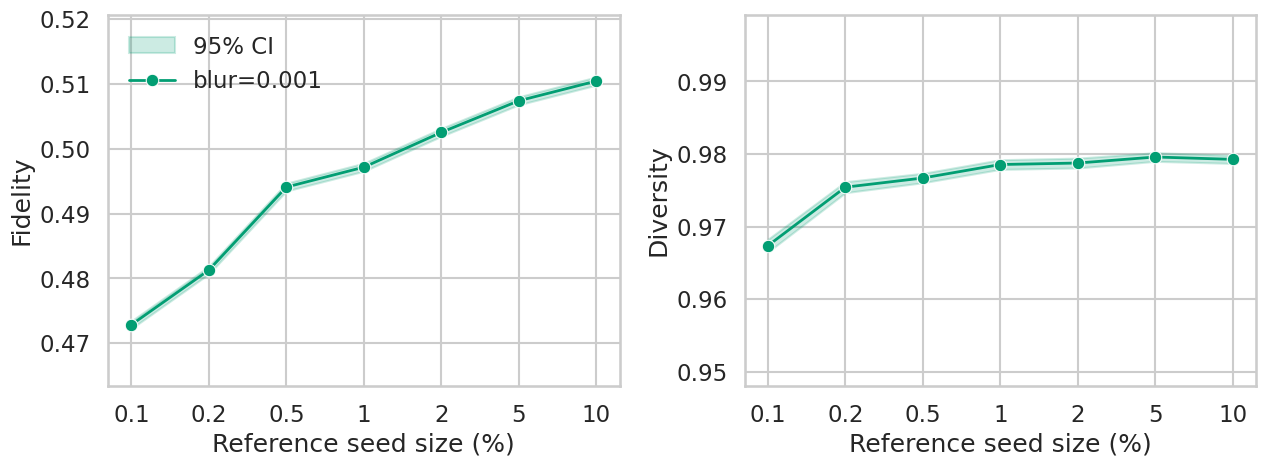}
  \caption{Fidelity + diversity of swapped GSM8K-style datasets.}
  \label{fig:gsm8k_swap}
\end{figure}

We also verify that our metrics capture fidelity even when approximated on empirical measures instead of mixing measures. To that end, we exactly replicate the setup used to produce Fig.~\ref{fig:gsm8k_metrics}, except we now swap the synthetic dataset being measured with the reference set. Instead of fixing the reference set, we now assess fidelity and diversity of the same dataset against increasingly diverse references. Consequently, we expect fidelity to improve monotonically as we increase the breadth of what we consider as the reference. This is in fact the observed outcome (Fig.~\ref{fig:gsm8k_swap}), suggesting that our metrics remain sensitive to differences in fidelity even on the level of empirical measures.

\subsection{SFT} \label{app:sft}
\begin{figure}[h]
\centering
\begin{subfigure}{.5\textwidth}
  \centering
  \includegraphics[width=.9\linewidth]{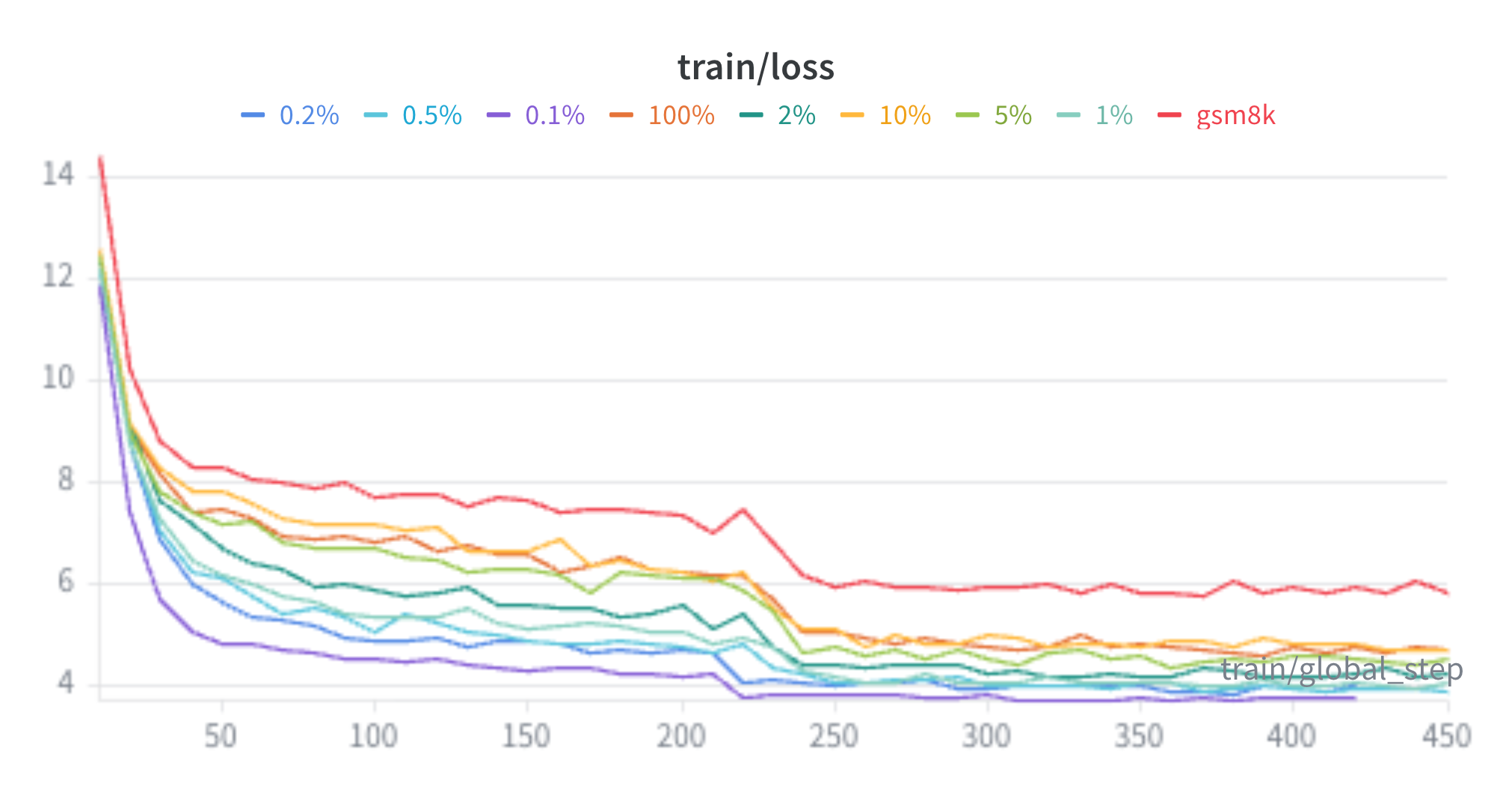}
  \caption{Llama-3.2-1B training loss.}
  \label{fig:1b_train}
\end{subfigure}%
\begin{subfigure}{.5\textwidth}
  \centering
  \includegraphics[width=.9\linewidth]{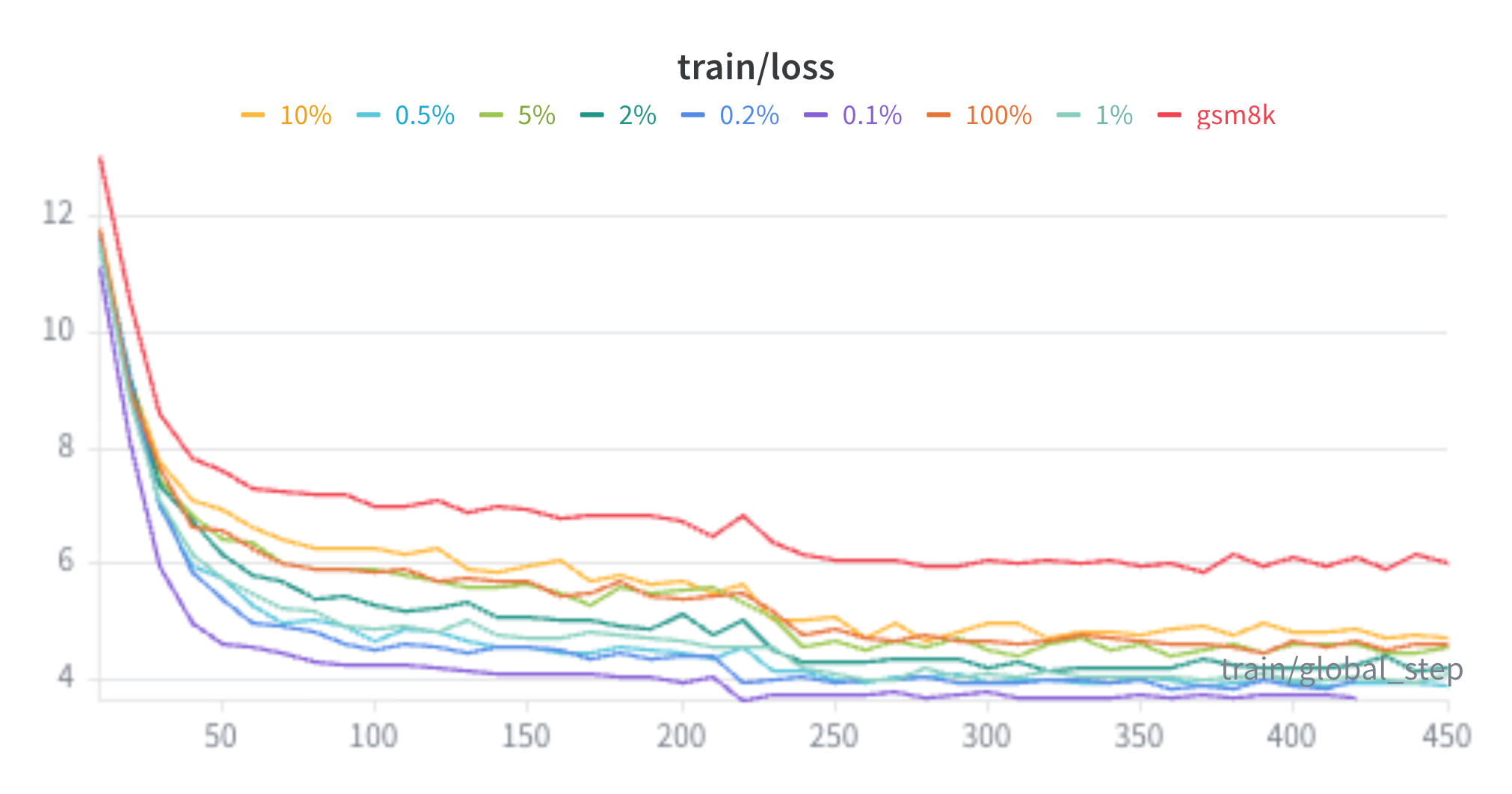}
  \caption{Llama-3.2-3B training loss.}
  \label{fig:3b_train}
\end{subfigure}

\begin{subfigure}{.5\textwidth}
  \centering
  \includegraphics[width=.9\linewidth]{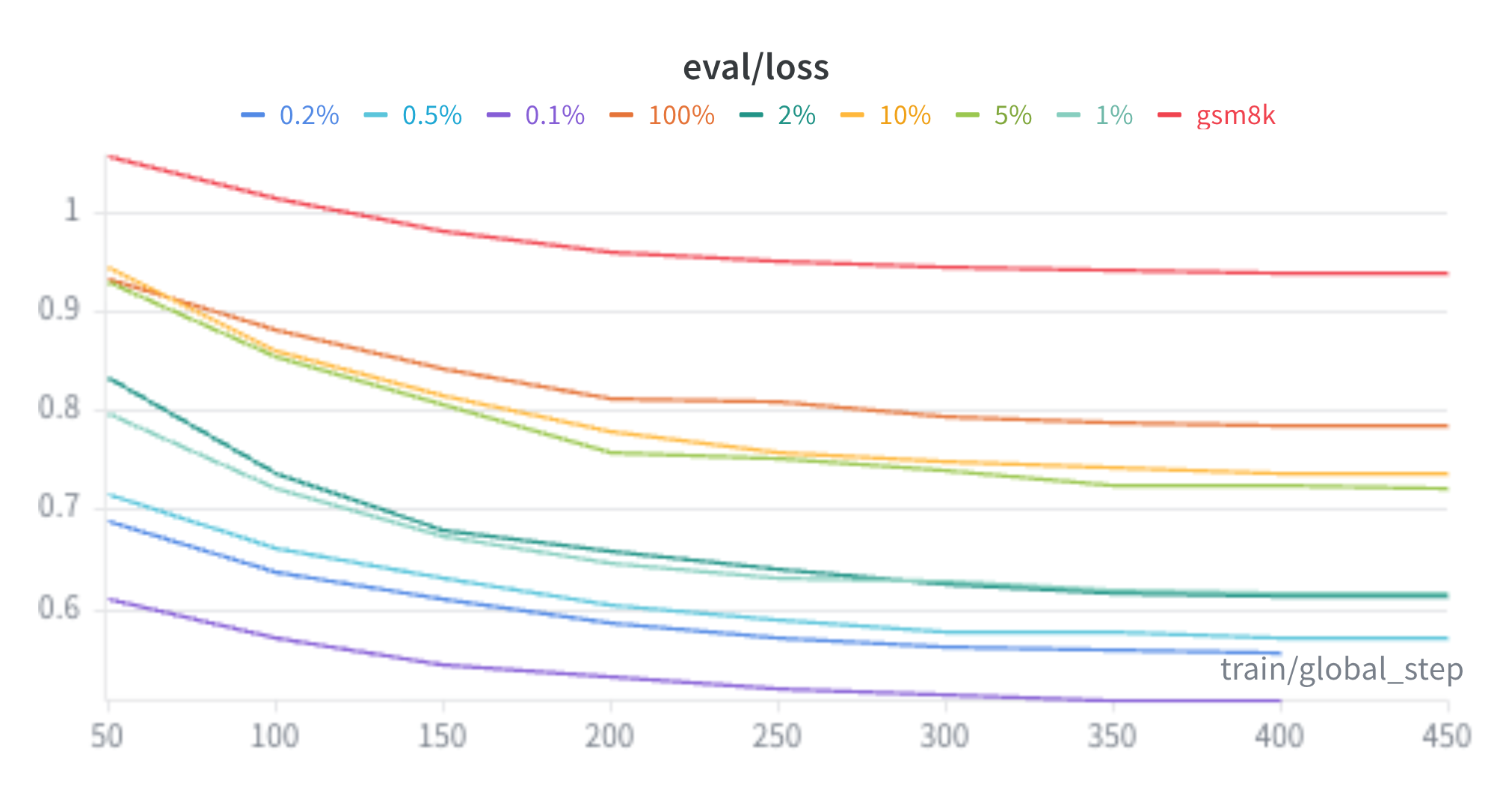}
  \caption{Llama-3.2-1B validation loss.}
  \label{fig:1b_val}
\end{subfigure}%
\begin{subfigure}{.5\textwidth}
  \centering
  \includegraphics[width=.9\linewidth]{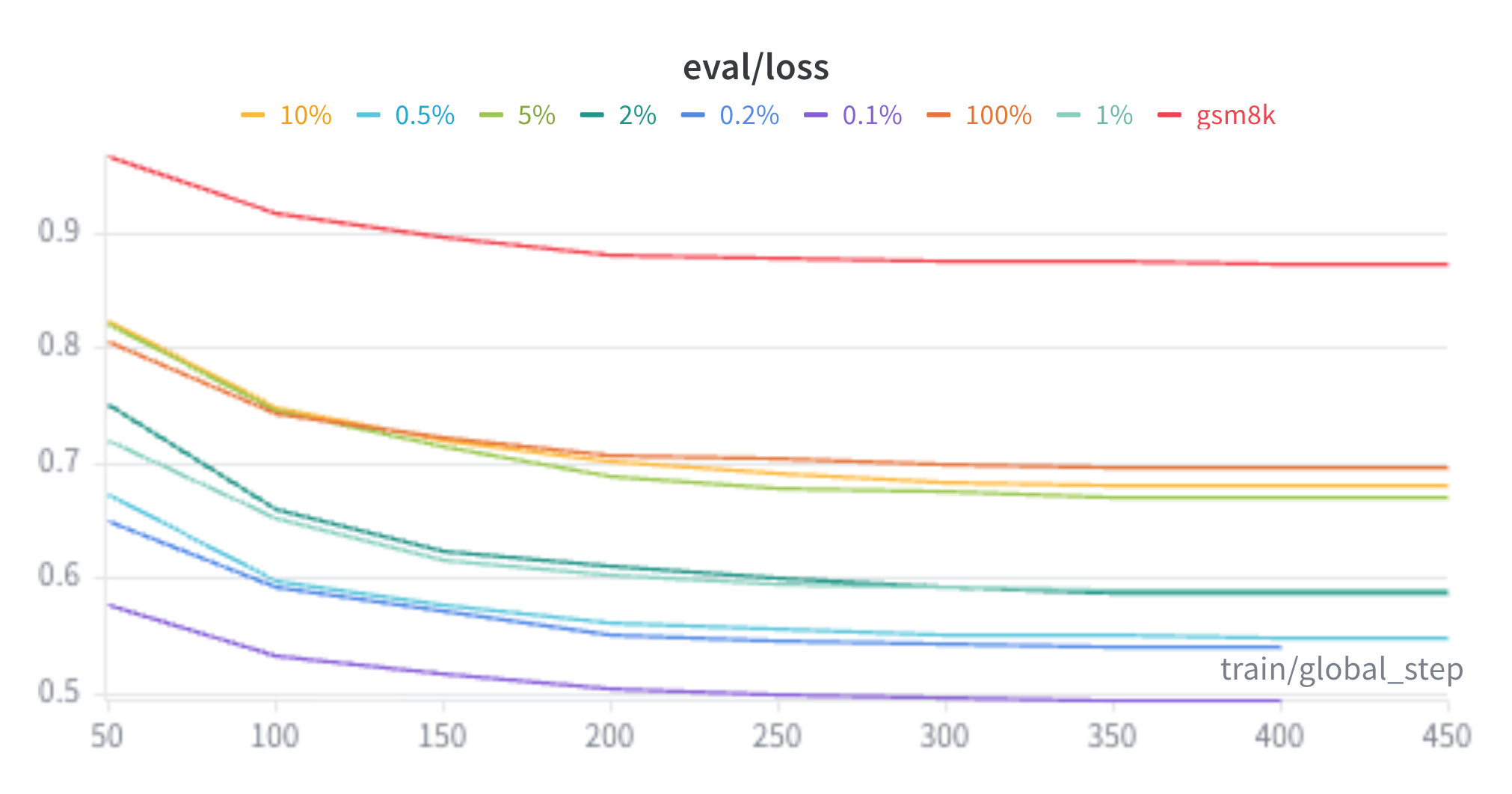}
  \caption{Llama-3.2-3B validation loss.}
  \label{fig:3b_val}
\end{subfigure}

\begin{subfigure}{.5\textwidth}
  \centering
  \includegraphics[width=.9\linewidth]{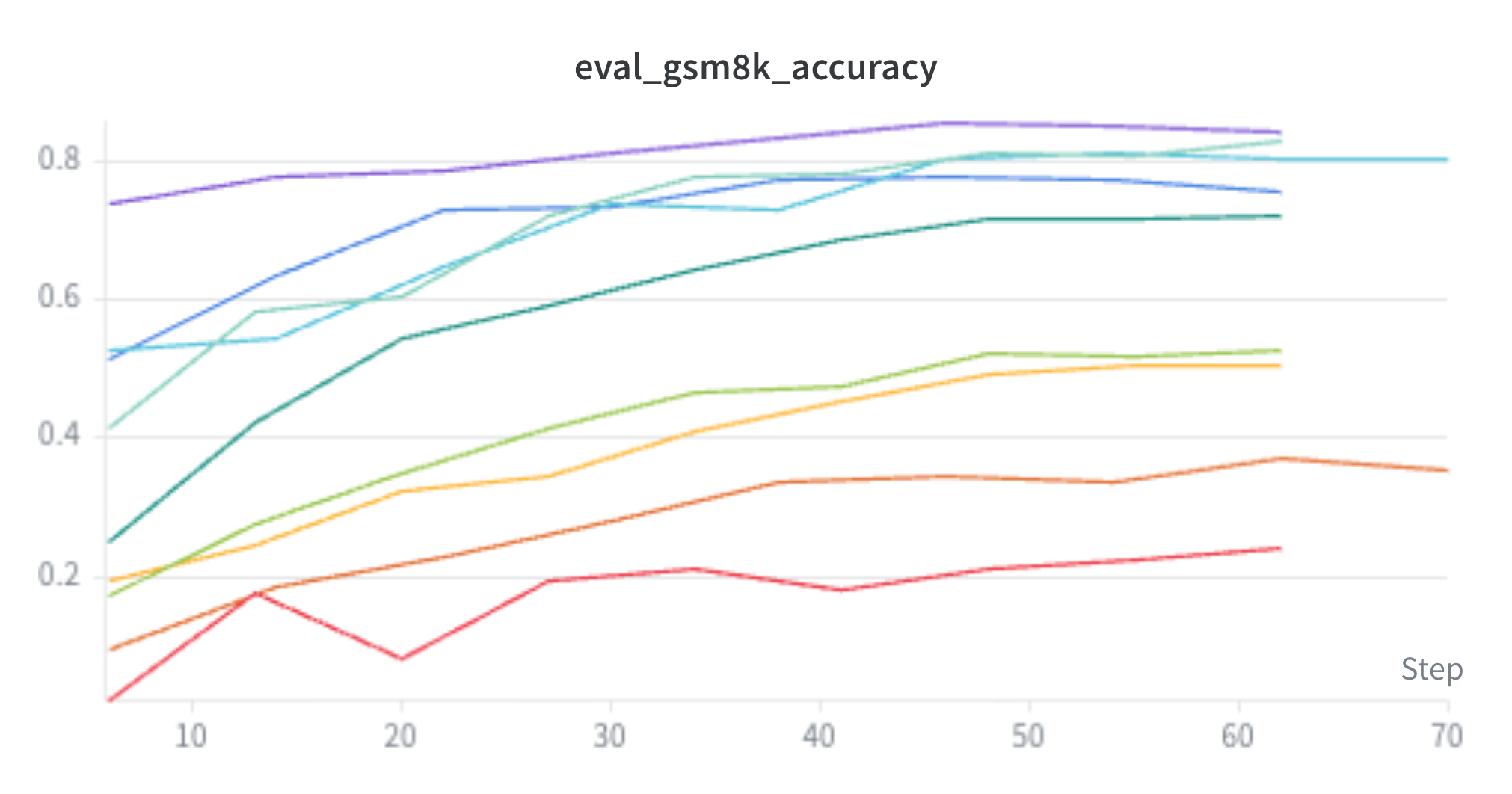}
  \caption{Llama-3.2-1B validation accuracy.}
  \label{fig:1b_acc}
\end{subfigure}%
\begin{subfigure}{.5\textwidth}
  \centering
  \includegraphics[width=.9\linewidth]{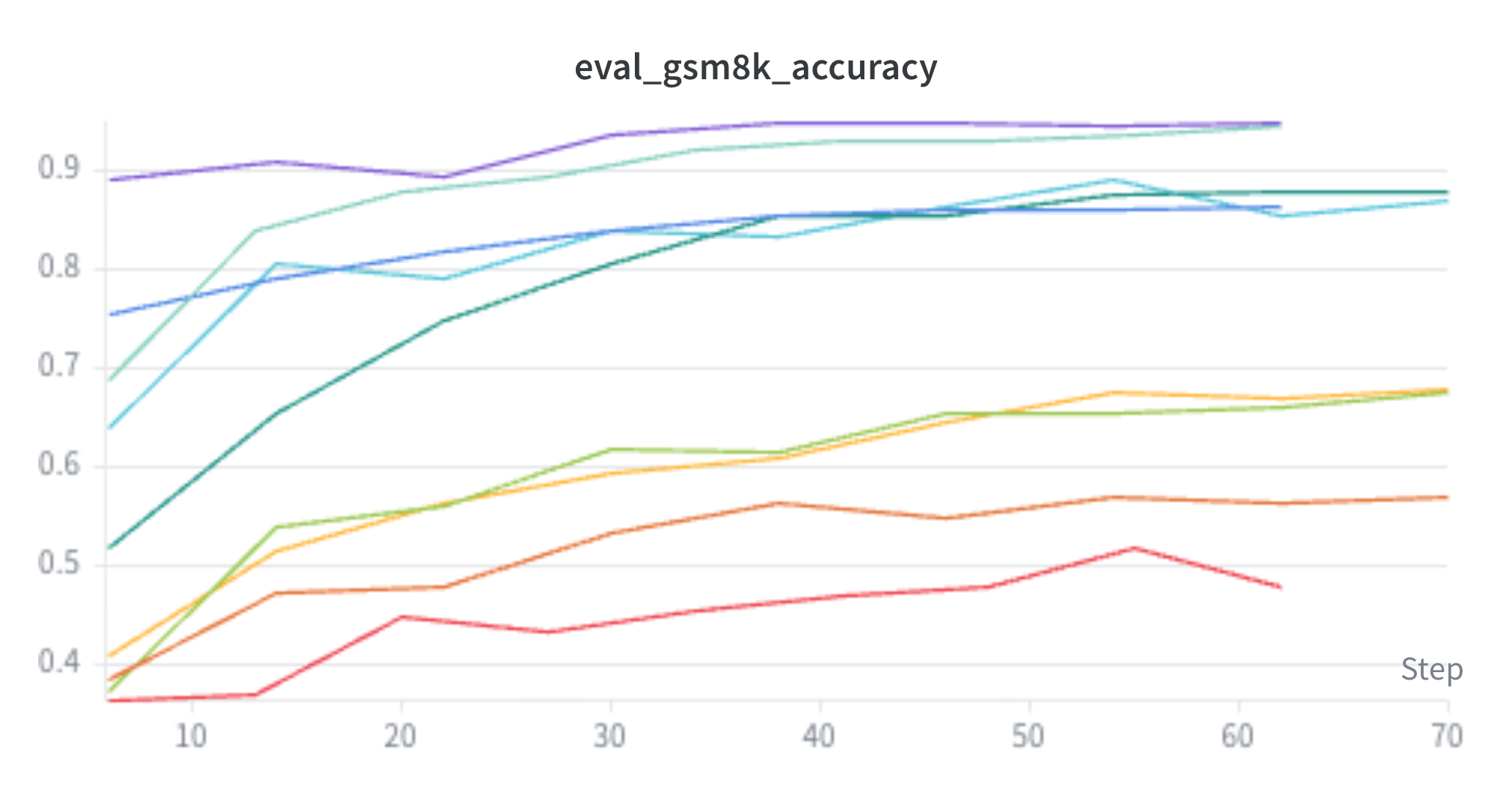}
  \caption{Llama-3.2-3B validation accuracy.}
  \label{fig:3b_acc}
\end{subfigure}
\caption{Training and validation curves from SFT on GSM8K.}
\label{fig:sft}
\end{figure}

We conduct supervised fine-tuning (SFT)on both Llama-3.2-1B and Llama-3.2-3B base models \cite{llama}. We finetuned a fresh copy of each base model on each of the eight synthetic datasets and the original GSM8K training set. All finetuning for both models was completed using the AdamW optimizer \cite{loshchilov2019decoupledweightdecayregularization} with cosine learning rate decay over a total of two epochs. We used a maximum learning rate of 4e-5 for the Llama-3.2-1B and 2e-5 for the Llama-3.2-3B. For all training datasets, we remove and set aside 200 randomly sampled examples to use for validation. Throughout training, we compute both cross-entropy loss and exact-match accuracy on the validation set to track progress. The final model used for evaluation on the GSM8K test set is loaded from the checkpoint achieving the highest validation accuracy. See Fig.~\ref{fig:sft} for all training and validation loss curves, as well as the validation accuracy used for model selection. All finetuning can be completed on an NVIDIA RTX A6000 GPU in 1-2 hours.

\subsection{Model evaluation}
\begin{figure}[H]
\centering
\begin{subfigure}{.5\textwidth}
  \centering
  \includegraphics[width=.8\linewidth]{gsm8k/llama-3.2-1b.png}
  \caption{Llama-3.2-1B}
  \label{fig:gsm8k_llama-1b}
\end{subfigure}%
\begin{subfigure}{.5\textwidth}
  \centering
  \includegraphics[width=.8\linewidth]{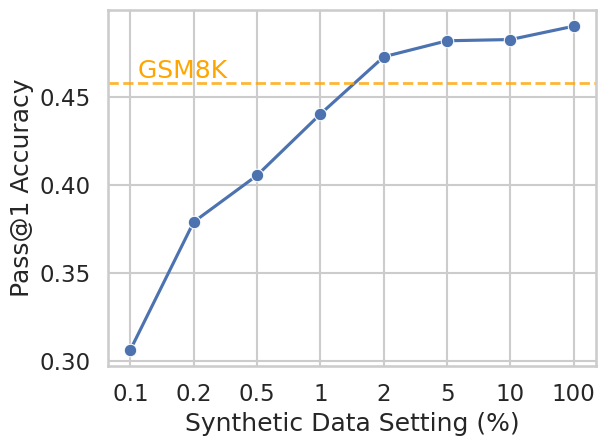}
  \caption{Llama-3.2-3B}
  \label{fig:gsm8k_llama-3b}
\end{subfigure}
\caption{Exact-match test accuracies of Llama-3.2 base models finetuned on synthetic datasets, evaluated on the GSM8K test split. Values on the x-axes indicate the percentage of original GSM8K training examples included in the seed data for that synthetic dataset. The dashed orange line indicates the test accuracy achieved by finetuning a base model on the entire original GSM8K training split.}
\label{fig:gsm8k_llama_test-acc}
\end{figure}

Each finetuned model is evaluated on exact-match accuracy on the GSM8K test set, with basic filtering applied to extract the final answer from the model generations. Test accuracies for both the 1B and 3B models are included in Fig.~\ref{fig:gsm8k_llama_test-acc}. We see a clear upward trend in test accuracies across both model sizes, with the larger model extracting strong test performance even from training on synthetic datasets of smaller  seed size.

\section{Proof of equation~\eqref{eq:nn_identity}}
\label{app:lemma}
In this Appendix, we provide a brief proof of the identity~\eqref{eq:nn_identity}, which
reduces to showing
\begin{equation} 
\inf_{\mu \in \calP(H)}W_1(G,\mu) =  \sum_{i=1}^k \alpha_i \cdot \min_{1 \leq j \leq k }\|\theta_i-\eta_j\|.
\end{equation}
The left-hand side
can be expressed as 
$$\inf_{\mu\in \calP(H)} W_1(G,\mu)
 = \inf_{\gamma\in \Pi(G)} \sum_{i,j=1}^k \gamma_{ij} \|\theta_i-\eta_j\|,$$
 where   $\Pi(G)$ denotes the set of all
 stochastic matrices $\gamma\in \bbR_+^{k\times k}$
 such that $\sum_{j=1}^k \gamma_{ij} = \alpha_i$
 for each $i=1,\dots,k$. Unlike the set of couplings $\Pi(G,H) $,
 notice that $\Pi(G)$ places no constraint on the second
 marginal of $\gamma$. 
 Now,  for any $\gamma \in \Pi(G)$, we have
$$\sum_{i,j=1}^k \gamma_{ij}\|\theta_i-\eta_j\|
\geq \sum_{i,j=1}^k \gamma_{ij} \min_{1 \leq j' \leq k} \|\theta_i-\eta_{j'}\|
=\sum_{i=1}^k \alpha_i \cdot \min_{1 \leq j \leq k} \|\theta_i-\eta_j\|.$$ 
Taking the infimum over all $\gamma \in \Pi(G)$ leads to 
 $$\inf_{\mu\in\calP(H)} W_1(G,\mu) \geq 
 \sum_{i=1}^k \alpha_i \cdot  \min_{1 \leq j \leq k}\|\theta_i-\eta_j\|.$$
 The lower bound is achieved by the coupling
$$\gamma_{ij} = \alpha_i \cdot I\Big(j = \argmin_{1\leq j'\leq k} \|\theta_i-\eta_{j'}\|\Big),\quad i,j=1,\dots,k,$$
breaking ties arbitrarily. The claim thus follows.


\end{document}